\documentclass[times, review, 10pt]{elsarticle}
\usepackage{amssymb}
\usepackage{graphicx}
\usepackage{amsmath}
\usepackage{longtable}
\usepackage{a4wide}
\usepackage{mathrsfs}
\usepackage{subcaption}
\usepackage{mathtools}
\usepackage{pifont}
\usepackage{algorithmic}
\usepackage{algorithm}
\usepackage{xcolor}
\usepackage{color}
\usepackage{lineno}
\usepackage{makecell} 
\usepackage{pdflscape}
\usepackage{adjustbox}
\usepackage[utf8]{inputenc}
\usepackage{tabularx}
\usepackage{setspace}
\usepackage{blindtext}
\usepackage{multirow}
\usepackage{notoccite} 
\usepackage{lscape} 
\usepackage{caption} 
\usepackage{mwe}
\usepackage{tikz}
\usepackage{siunitx}
\usepackage{mathrsfs}
\usetikzlibrary{shapes,arrows}
\usepackage{xcolor}
\usepackage{booktabs}
\usepackage{hyperref}
\usepackage{amsthm}

\newcommand{\RNum}[1]{\lowercase\expandafter{\romannumeral #1\relax}}
\newcommand{\RNumU}[1]{\uppercase\expandafter{\romannumeral #1\relax}}
\usepackage{natbib}
\journal{Elsevier}

\begin{document}
\begin{frontmatter}
\title{GB-RVFL: Fusion of Randomized Neural Network and Granular Ball Computing}
\author[inst1]{M. Sajid}
\ead{phd2101241003@iiti.ac.in}
\author[inst1]{A. Quadir}
\ead{mscphd2207141002@iiti.ac.in}
\author[inst1]{M. Tanveer\corref{Correspondingauthor}}
\ead{mtanveer@iiti.ac.in}
\author[]{for the Alzheimer’s Disease Neuroimaging Initiative\corref{authorADNI}}

\affiliation[inst1]{organization={Department of Mathematics, Indian Institute of Technology Indore},
            addressline={Simrol}, 
            city={Indore},
            postcode={453552}, 
            state={Madhya Pradesh},
            country={India}}
            \cortext[Correspondingauthor]{Corresponding author}
            \cortext[authorADNI]{This study used data from the Alzheimer's Disease Neuroimaging Initiative (ADNI) (\url{adni.loni.usc.edu}). The ADNI investigators were responsible for the design and implementation of the dataset, but they did not take part in the analysis or the writing of this publication.  \url{http://adni.loni.usc.edu/wp-content/uploads/how\_to\_apply/ADNI\_Acknowledgement\_List.pdf} has a thorough list of ADNI investigators.}
\begin{abstract}
The random vector functional link (RVFL) network is a prominent classification model with strong generalization ability.  However, RVFL treats all samples uniformly, ignoring whether they are pure or noisy, and its scalability is limited due to the need for inverting the entire training matrix. To address these issues, we propose granular ball RVFL (GB-RVFL) model, which uses granular balls (GBs) as inputs instead of training samples. This approach enhances scalability by requiring only the inverse of the GB center matrix and improves robustness against noise and outliers through the coarse granularity of GBs. Furthermore, RVFL overlooks the dataset's geometric structure. To address this, we propose graph embedding GB-RVFL (GE-GB-RVFL) model, which fuses granular computing and graph embedding (GE) to preserve the topological structure of GBs. The proposed GB-RVFL and GE-GB-RVFL models are evaluated on KEEL, UCI, NDC and biomedical datasets, demonstrating superior performance compared to baseline models.
\end{abstract}

\end{frontmatter}
\section{Introduction}
The randomization-based neural networks (RNNs) \cite{schmidt1992feed} have been effectively used for a wide range of classification and regression tasks due to their universal approximation capabilities \cite{igelnik1995stochastic, huang2006extreme}. Generally, the backpropagation (BP)-based algorithm is extensively employed to train feedforward neural networks (NNs). However, BP-based algorithms come with many underlying issues such as potential slowness, susceptibility to local optima \cite{gori1992problem}, and the critical influence of factors such as learning rate and initialization point. RNNs emerged as a solution to the drawbacks of BP-based NNs mentioned earlier. In RNNs, some parameters are initialized and remain fixed throughout the training process \cite{sajid2024intuitionistic}, while the parameters of the output layer are determined through a closed-form solution.

The random vector functional link (RVFL) network  \cite{pao1994learning, malik2023random} is a shallow feed-forward RNN characterized by randomly initialized hidden layer parameters that remain untouched throughout the training process. RVFL stands out among other RNNs due to its direct connections between input and output layers. These direct links act as a form of implicit regularization \cite{zhang2016comprehensive} within RVFL, leading to improved learning capabilities. By employing methods like the Pseudo-inverse or least-squares technique, RVFL delivers a closed-form solution for optimizing output parameters, resulting in efficient learning with fewer adjustable parameters. In addition to that, RVFL demonstrates universal approximation capability \cite{igelnik1995stochastic}.

However, in the closed-form solution, RVFL involves matrix inverse computation of the whole training matrix (see subsection \ref{RVFL_Section}), which may be intractable in large-scale problems. Additionally, in the conventional RVFL, each sample receives an equal weighting during the creation of the optimal classifier, leaving it susceptible to noise despite its robust generalization in clean datasets. To address this issue, fuzzy theory has proven effective in reducing the negative impact of noise or outliers on the performance of machine learning models \cite{rezvani2019intuitionistic, quadir2024intuitionistic}. Intuitionistic fuzzy (IF) is an extended version of fuzzy concepts, which uses membership and nonmembership functions to give an IF score to each sample. In \cite{malik2022alzheimer, ganaie2024graph}, intuitionistic fuzzy RVFL (IFRVFL) and graph embedding IFRVFL for class imbalance learning (GE-IFRVFL-CIL) were proposed with the 
aim to address the challenges posed by noise and outliers in datasets.  However, these models have two associated challenges. Firstly, they require the computation of membership and non-membership values in the kernel space, which increases the computational complexity of IFRVFL and GE-IFRVFL-CIL. Secondly, IFRVFL and GE-IFRVFL-CIL also require the computation of inverses for the whole training sample matrix while calculating the output layer parameters, which makes it unsuitable for large-scale datasets.

Moreover, the traditional RVFL overlooks the geometric aspects of the data when determining the final output parameters \cite{malik2023random}. Several enhanced variants of the RVFL have emerged as solutions to this issue \cite{ganaie2024graph, MalikGraph2022}. However, the developed models also need to use whole training datasets in the calculation of the inverse matrix in the closed-form solution. 

Human cognition prioritizes a ``large scope first" principle, with the visual system emphasizing global topological features, processing information from larger to smaller scales. Inspired by this, \citet{xia2019granular} developed a classifier using GBs, leveraging GBs to categorize datasets based on different granular sizes \cite{xia2020fast}. Larger granularity sizes align with scalable and efficient approaches, resembling human cognitive processes \cite{xia2021granular}. However, shifting toward larger granularity may sacrifice detail and accuracy, whereas finer granularity enhances detail focus but may compromise scalability. Achieving a balanced granular size is crucial. Scholars have extensively researched breaking down information into different granularities for various tasks \cite{zhang2021double,qin2023overview}, enhancing the effectiveness of multi-granularity learning in addressing real-world challenges \cite{pedrycz1984identification, xie2024adaptive}.

To address noise and outliers present in the dataset, an approach integrating the concept of GBs into support vector machine (SVM), namely, granular ball SVM (GBSVM) \cite{xia2022gbsvm}, was proposed. GBSVM utilizes GBs derived from the dataset as inputs instead of the conventional use of individual data points. GBSVM has demonstrated proficiency in handling noise and outliers, showcasing better scalability when compared to the standard SVM approach.

Inspired by the effectiveness and scalability of a granular approach in handling noise and outliers, in this paper, we fuse it with the RVFL and propose the granular ball RVFL (GB-RVFL). The proposed GB-RVFL leverages GBs as inputs and need to inverse the matrix of centers of the GBs rather than the matrix of the whole training dataset, resulting in improved scalability and a heightened ability to withstand noise and outliers. Additionally, to maintain the intrinsic geometric structure within the dataset, we integrate graph embedding (GE) \cite{yan2006graph} into GB-RVFL, resulting in the proposed graph embedded GB-RVFL (GE-GB-RVFL). GE-GB-RVFL offers several advantages: $(i)$ by operating on the inverse of the matrix of GB centers, it enhances scalability and is well-suited for large-scale data compared to RVFL. $(ii)$ Leveraging granularity concepts, GE-GB-RVFL effectively mitigates the adverse effects of noise and outliers. $(iii)$ The GE framework preserves the intrinsic topological arrangement of datasets, providing GE-GB-RVFL with the advantage of utilizing the inherent dataset structure, thereby increasing its efficacy. To the best of our knowledge, this marks the inaugural instance where the RVFL model incorporates GB as an input rather than individual point samples.\\
 The paper's key highlights are as follows:
\begin{enumerate}
    \item We propose the GB-RVFL model, which uses GBs as input rather than individual input samples for classifier construction. The utilization of GBs enhances the scalability of the proposed GB-RVFL by requiring only the inverse of the GB centers matrix rather than the matrix of entire samples. Additionally, this design improves the proposed GB-RVFL model's robustness against noise and outliers using the coarse granularity of GBs.
    \item Further, we propose the GE-GB-RVFL model, aiming to preserve the dataset's intrinsic geometric structure while retaining the GB-RVFL model's core properties. This model incorporates subspace learning (SL) criteria for output weight computation within the GE framework (integrating intrinsic and penalty SL). The incorporation of a graph regularization term in conjunction with GE serves the purpose of preserving the structural details of the graph in the projection space.
    \item The performance evaluation of the proposed GB-RVFL and GE-GB-RVFL models involves testing on benchmark KEEL and UCI datasets. These datasets are sourced from various domains and exhibiting different sizes, and are tested with and without label noise, comparing the performance of our proposed models against existing ones. Additionally, experiments on NDC datasets to demonstrate the effectiveness of the proposed model for large data. 
    \item To demonstrate the practical applicability of the proposed GB-RVFL and GE-GB-RVFL models, we apply them to real-world biomedical datasets, specifically the BreakHis dataset for breast cancer classification and the ADNI datasets for the classification of Alzheimer's disease.
    \item Finally, we demonstrate the enhanced feature interpretability of the proposed GB-RVFL and GE-GB-RVFL models.
\end{enumerate}
The succeeding sections of this paper are structured as follows: Section \ref{Related_works} introduces GB Computing, RVFL, and GE. Section \ref{proposed_work} details the mathematical framework of the proposed GB-RVFL and GE-GB-RVFL models. Experimental results and analyses of proposed and existing models are discussed in Section \ref{experiments}. In Section \ref{interpretability}, we show the enhancement in the feature interpretability of the proposed models. In Section \ref{discussion}, we engage in discussions grounded in empirical evaluations. Conclusion and some potential future research directions are outlined in Section \ref{conclusion}.
\section{Related Works}
\label{Related_works}
In this section, we first define some notations and then discuss granular computation, the mathematical framework of RVFL, and graph embedding (GE).
\subsection{Notations}
Let $M$ be the total number of training samples and the training set is $T = \bigl\{(v_i,z_i)\vert\, v_i \in \mathbb{R}^{1 \times P},\, z_i \in \mathbb{R}^{1 \times C}, ~i = 1, 2, \cdots, M\bigl\}$. Let $V=\bigl(v_1^t,v_2^t,\hdots,v_M^t\bigl)^t \in \mathbb{R}^{M \times P}$ and $Z=\bigl(z_1^t,z_2^t,\hdots,z_M^t\bigl)^t \in \mathbb{R}^{M \times C}$ be the collection of all input and target vectors, respectively, where $(\cdot)^t$ is the transpose operator. $g$ denotes the number of hidden layer nodes. Let $k$ number of GBs generated on $T$ be $\{GB_1, GB_2, \dots, GB_k\}$ and $o_j$ be the center of the granular ball $GB_j$ for $j=1,2,\hdots,k$. $\otimes$ and $\oplus$ denote the  Kronecker product and the Concatenation operator, respectively and are defined below. For $C \in \mathbb{R}^{r \times s}$, $D \in \mathbb{R}^{t \times u}$ and $E \in \mathbb{R}^{t \times v}$, then \\
$C \otimes D = \begin{bmatrix}
c_{11}D&\hdots&c_{1s}D \\
\vdots & \vdots  & \vdots \\
c_{r1}D&\hdots&c_{rs}D
 \end{bmatrix} \in \mathbb{R}^{rt \times su}$ and $D \oplus E = [D~ E] \in \mathbb{R}^{t \times (u+v)}.$

\subsection{Granular Ball Computation \cite{xia2019granular}}
In 1996, Lin and Zadeh introduced the concept of ``granular computing" with the goal of minimizing the quantity of training data points. This approach aims to capture the essence of data simplification while preserving representativeness during the learning process. Consider a granular ball (GB) encompassing $q$ data points, i.e., $\{v_1, v_2, \dots, v_q\}$, where each $v_j$ belongs to the vector space $\mathbb{R}^{1\times P}$. The center $o$ of a $GB$ is defined as the center of gravity calculated from all sample points within the ball. Mathematically, they can be calculated as: $o = \frac{1}{q} \sum_{j=1}^{q}v_j$.

The class/label assigned to a GB is based on the labels of the data points with the highest occurrence of data samples within the ball. To quantitatively evaluate the degree of division within a GB, the notion of ``threshold purity" is introduced. This threshold purity signifies the proportion of the predominant samples sharing the same label within the GB. The optimization problem governing the generation of GBs on set $T$ is expressed as follows:
\begin{align}
    \text{min}\hspace{0.2cm}&\gamma_1 \times \frac{M}{\sum_{j=1}^{k}|GB_j|} + \gamma_2 \times k ,\nonumber\\
        \text { s.t. } \hspace{0.1cm} & purity(GB_j) \geq \rho, ~~ j=1,2,\hdots,k,
    \end{align}
where $\gamma_1$ and $\gamma_2$ are weight coefficients. $\rho$ is the threshold purity and $|\cdot|$ represents the cardinality of a GB. 
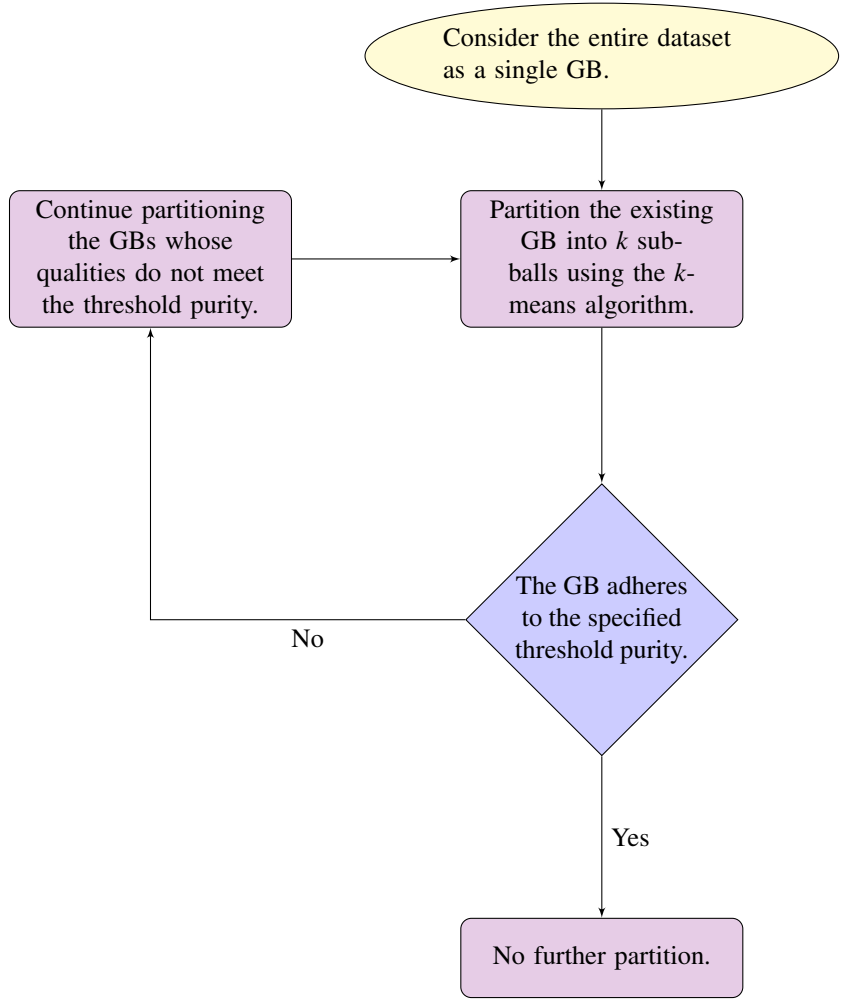
\begin{figure}[ht!]
    \centering       
\tikzstyle{decision} = [diamond, draw, fill=blue!20, 
    text width=7.0em, text badly centered, node distance=1.2cm, inner sep=0pt]
\tikzstyle{block} = [rectangle, draw, fill=violet!20, 
    text width=10em, text centered, rounded corners, minimum height=3em]
\tikzstyle{line} = [draw, -latex']
\tikzstyle{cloud} = [draw, ellipse,fill=yellow!20, node distance=2cm,
    minimum height=4em,text width=12.0em]
\begin{tikzpicture}[node distance =2.7cm, auto]
    \node [cloud] (init) {Consider the entire dataset as a single GB.};
    \node [block, below of=init] (identify) {Partition the existing GB into $k$ sub-balls using the $k$-means algorithm.};
    \node [block, left of=identify, node distance=6.0cm] (update) {Continue partitioning the GBs whose qualities do not meet the threshold purity.};
    \node [decision, below of=identify,node distance=4.8cm] (decide) {The GB adheres to the specified threshold purity.};
    \node [block, below of=decide, node distance=4.5cm] (stop) {No further partition.};
    \path [line] (init) -- (identify);
    \path [line] (identify) -- (decide);
    \path [line] (decide) -| node [near start] {No} (update);
    \path [line] (update) -- (identify);
    \path [line] (decide) -- node {Yes}(stop);
\end{tikzpicture}
\caption{Granular ball generation process.}
    \label{Process of the granular-ball generation}
\end{figure}
The entire dataset is treated as a unified GB at the outset. When the purity of the GB falls below the given threshold, it must be divided several times until all sub-GBs reach or exceed the threshold purity. As the GBs' purity improves, so does their alignment with the original dataset's data distribution. Figure \ref{Process of the granular-ball generation} demonstrates the technique for generating GBs.

\begin{figure*}[ht]
\begin{minipage}{.333\linewidth}
\centering
\subfloat[The original dataset.]{\label{fig:1a}\includegraphics[scale=0.42]{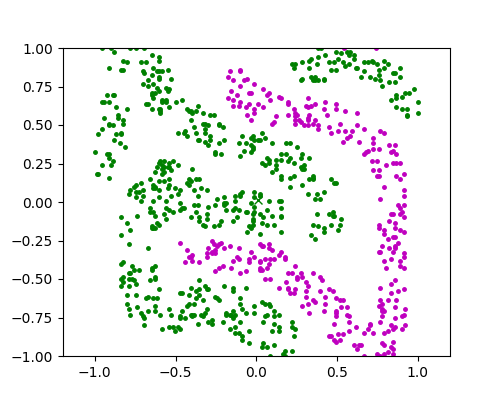}}
\end{minipage}
\begin{minipage}{.333\linewidth}
\centering
\subfloat[Generated granular balls in the first iteration]{\label{fig:1b}\includegraphics[scale=0.42]{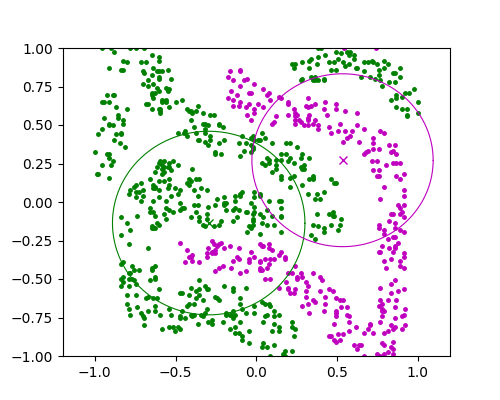}}
\end{minipage}
\begin{minipage}{.333\linewidth}
\centering
\subfloat[Generated granular balls in the second iteration]{\label{fig:1c}\includegraphics[scale=0.42]{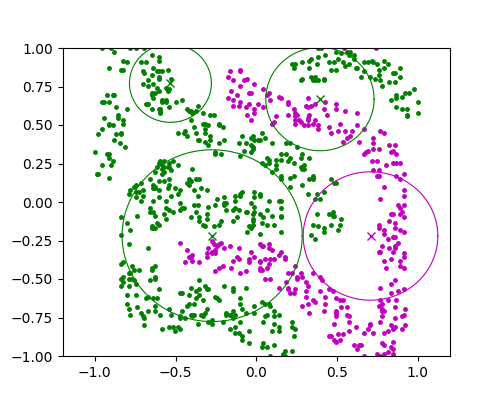}}
\end{minipage}
\par\medskip
\begin{minipage}{.333\linewidth}
\centering
\subfloat[Generated granular balls in the middle iteration]{\label{fig:1d}\includegraphics[scale=0.42]{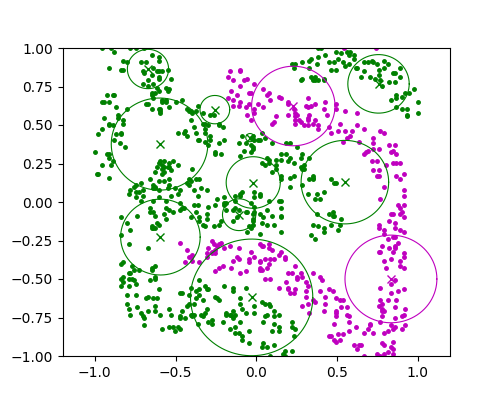}}
\end{minipage}
\begin{minipage}{.333\linewidth}
\centering
\subfloat[Results after stop splitting]{\label{fig:1e}\includegraphics[scale=0.42]{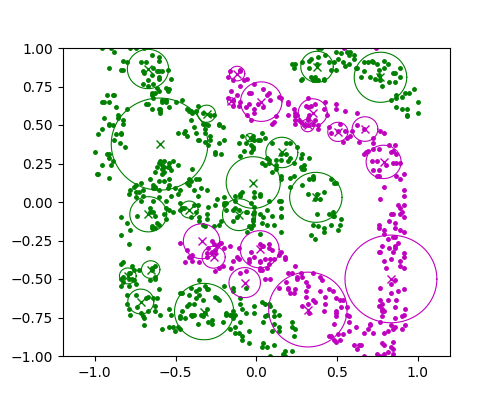}}
\end{minipage}
\begin{minipage}{.333\linewidth}
\centering
\subfloat[Extracted granular balls]{\label{fig:1f}\includegraphics[scale=0.42]{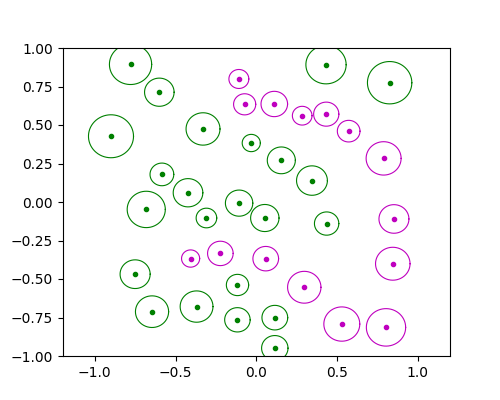}}
\end{minipage}
\caption{The visualization of generating granular balls by splitting the ``fourclass'' dataset. It assigns the label ``$+1$'' to the green granular balls, while the label ``$-1$'' is used for the magenta granular balls.}
\label{gb-generation}
\end{figure*}

In Fig. \ref{gb-generation}, the GB generation and splitting process using the ``fourclass" dataset is demonstrated. Initially, in Fig. \ref{fig:1a}, the dataset consists of two non-linearly separable classes, with green points labelled as ``$+1$'' and magenta points labelled as ``$-1$''. In the first iteration, shown in Fig. \ref{fig:1b}, large GBs are formed, loosely covering the green and magenta data points, providing a rough approximation of the data structure. As the process continues in Fig. \ref{fig:1c} and Fig. \ref{fig:1d}, the GBs are progressively refined through iterative splitting, becoming smaller and more localized, capturing finer details of the dataset. By the final stages, shown in Fig. \ref{fig:1e} and Fig. \ref{fig:1f}, the GBs effectively encapsulate the underlying structure of the two classes, creating well-defined boundaries that can be used for classification purposes.

\subsection{Random Vector Functional Link (RVFL) Network \cite{pao1994learning}}
\label{RVFL_Section}
\begin{figure}[h]
\centering
\includegraphics[width=0.62\textwidth]{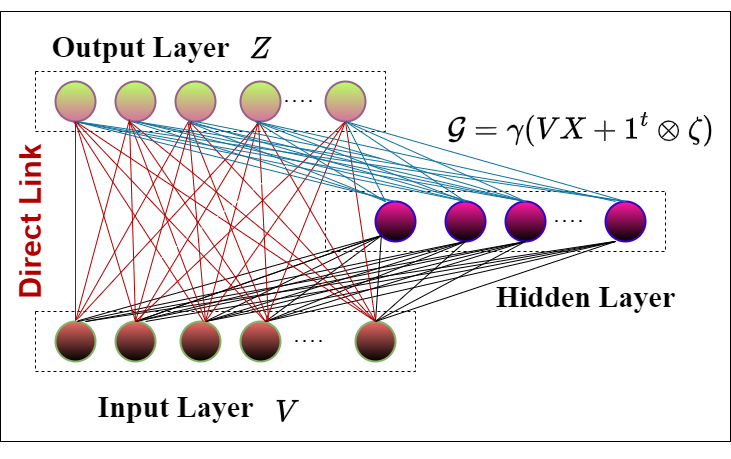}
\caption{Visual representation depicting the framework of the RVFL.}
\label{fig:RVFL}
\end{figure}

The RVFL, proposed by \cite{pao1994learning}, comprises input and output layers along with a single hidden layer. Throughout the training phase, the biases and weights of the hidden layer are initialized at random from uniform distributions within the domains [0,1] and [-1,1], respectively, and remain fixed. The hidden layer and the input layer (facilitated by direct links) are connected to the output layer through the output layer weights. The calculation of output layer weights employs analytical methods such as the least square technique or Pseudo inverse. Refer to Figure \ref{fig:RVFL} for an illustration of the RVFL model's architecture.
\label{proposed_work}
 
Consider $\mathcal{G}$ as the hidden layer matrix, obtained by randomly projecting the input matrix and then applying the non-linear activation function $\gamma$, which is defined as:
 \begin{align}
 \label{G_1_layer}
        \mathcal{G}=\gamma(V X + \mathbf{1}^t \otimes \zeta) \in \mathbb{R}^{M \times g},
     \end{align}
where $X \in \mathbb{R}^{P \times g} $ is the randomly initialized weights matrix, $\mathbf{1}$ is a vector of ones, and $\zeta$ $\in \mathbb{R}^{1 \times g}$ is the bias vector. Therefore, $\mathcal{G}$ is given as:
\begin{align*}  
{\mathcal{G}}=\left[\begin{array}{ccc}
\gamma\left(v_1x_1+\zeta^{(1)}\right) & \ldots & \gamma\left( v_1x_{g}+\zeta^{(g)}\right) \\
\vdots & \vdots & \vdots \\
\gamma\left(v_Mx_1+\zeta^{(1)}\right) & \ldots & \gamma\left(v_M x_{g}+\zeta^{(g)}\right)
\end{array}\right],
\end{align*}

where $\zeta^{(j)}$ represents the $j^{th}$ hidden node's bias term, $x_k \in \mathbb{R}^{P \times 1}$ denotes the $k^{th}$ column vector of the weights connecting the $k^{th}$ node of the hidden layer to all the input nodes and $v_i \in \mathbb{R}^{1 \times P}$ is the $i^{th}$ row (original input) of the inputs matrix $V$. The output layer's weights are calculated as follows:
\begin{equation}\label{eq:train1}
[V \oplus \mathcal{G}]\mathcal{Q}=\hat{Z},    
\end{equation}
where $\mathcal{Q} \in \mathbb{R}^{(P + g) \times C}$ is the unknown weights matrix that connects the output layer with the input layer and hidden layers. $\hat{Z}$ is the anticipated output.
$\mathcal{Q}$ can be calculated as:
\begin{align}
    \mathcal{Q}&=[V \oplus \mathcal{G}]^{-1}Z, \label{eq:inverse}\\~\text{or,}~\mathcal{Q}&=[V \oplus \mathcal{G}]^{\dagger}Z, \label{eq:pseudo-inverse}
\end{align}
where $[\cdot]^{-1}$ and $[\cdot]^{\dagger}$ denote the inverse and pseudo-inverse, respectively. If $[V \oplus \mathcal{G}]$ is non-sigular, we use \eqref{eq:inverse} to calculate $\mathcal{Q}$, otherwise \eqref{eq:pseudo-inverse}. 

The inverse calculation (in \eqref{eq:inverse} and \eqref{eq:pseudo-inverse}) can be a significant challenge, particularly in scenarios involving large matrices. This task is computationally demanding and becomes a bottleneck for RVFL-based approaches when dealing with high-order matrices. Essentially, the complexity, memory usage, performance impact, and scalability concerns associated with finding the inverse of such matrices can hinder the performance of RVFL-based methods in real-world scenarios characterized by high-dimensional data and large sample sizes.
\subsection{Graph Embedding (GE) \cite{4016549}}
\label{subsection:G_matrix_Graph}
The GE process \cite{4016549} is designed to retain crucial graph structural details within the projection space. In the GE framework, considering an input dataset denoted as $V$, two components are defined: the intrinsic graph $\mathcal{U}^{int}=\{V,\Theta^{int}\}$ and the penalty graph $\mathcal{U}^{pen}=\{V,\Theta^{pen}\}$. The similarity weight matrix $\Theta^{int} \in \mathbb{R}^{M \times M}$ incorporates the graph weights corresponding to the intrinsic connections among vertices in $V$. Additionally, each element of the $\Theta^{pen} \in \mathbb{R}^{M\times M}$ represents the penalty matrix of $V$, which accounts for specific relationships among the graph's vertices. The optimization problem for GE is formulated as follows:
\begin{align} 
\label{eqn:GE1}
\hat{y}=&\underset {Tr({y_{0}^{t}V^{t}\mathcal{S}V{y_{0}})} = a}{\rm argmin} \sum _{k\neq l} \left \|{  {y_{0}}^{t}  {v}_{k} -  {y_{0}}^{t}  {v}_{l} }\right \|_2^{2} \Theta_{kl}^{int}, \notag \\=&\underset {Tr({ {y_{0}}^{t}  {V}^{t}  \mathcal{S}  {V}  {y_{0}})}= a}{\rm argmin} {  Tr({y_{0}}^{t}  {V}^{t}  {\mathcal{L}} {V} {y_{0}}) }. \end{align}
Here, $y_{0}$ represents the projection matrix, and the trace operator is denoted as $Tr(\cdot)$.

For $\mathcal{U}^{int}$, the Laplacian matrix is represented by $\mathcal{L}=\mathcal{F}-\Theta^{int}\in \mathbb{R}^{M \times M}$, where $\mathcal{F}$ is the diagonal matrix, whose diagonal elements are defined as $\mathcal{F}_{kk}=\sum_{l}\Theta_{kl}^{int}$. $\mathcal{S}=\mathcal{L}^{pen}=\mathcal{F}^{pen}-\Theta^{pen}$ is the Laplacian matrix of $\mathcal{U}^{pen}$, and $a$ is a scalar term. The Eq. (\ref{eqn:GE1}) reduces to a generalized eigenvalue problem \cite{chung1997spectral} as follows:
\begin{align}
  {U}_{int} {s} = \lambda  {U}_{pen} {s},\end{align}
here $U_{int}=V^{T}\mathcal{L}V$ and ${U}_{pen}=V^{T}\mathcal{S}V$. This indicates that the eigenvectors of the matrix $U={U}_{pen}^{-1}{U}_{int}$ will be used to generate the transformation matrix. The matrix $U$ encompasses the inherent relationships among data samples through the intrinsic and penalty graph connections.
\section{Proposed Work}
\label{proposed_work}
In this section, we first give the formulation of the proposed GB-RVFL, and then we discuss detailed mathematical formulation along with the solution of the proposed GE-GB-RVFL model. Let $\{GB_1, GB_2, \dots, GB_k\}$ be the set of GBs for the training dataset $T$. Let $o_j$ and $w_j$ be the center and label of the granular ball $GB_j$, respectively. Let $\mathcal{O}=[o_1^t,o_2^t,\hdots,o_k^t]^t \in \mathbb{R}^{k \times P}$ and $\mathcal{W}=[w_1^t,w_2^t,\hdots,w_k^t]^t \in \mathbb{R}^{k \times C}$ be the matrix of centers and classes of GBs, respectively.

\subsection{Granular Ball Random Vector Functional Link (GB-RVFL) Network}
Through the fusion of RVFL and GB computing, we propose GB-RVFL with the aim of achieving greater scalability and robustness compared to the RVFL model. We first highlight the rationale behind the scalability and robustness of our model, and then we delve into the mathematical framework that fuses GB and RVFL to form GB-RVFL.
\begin{itemize}
    \item \textbf{\textit{Robustness:}} Our proposed GB-RVFL model fuses the concept of GBs with RVFL. During the training, the proposed GB-RVFL captures information from either the entire sample space or from subsets of the sample space (in the form of GBs). These GBs, derived from the training dataset, are inherently coarse and represent a small fraction of the total data points. By leveraging the coarse nature of GBs, specifically focusing on their centers, we effectively harness the bulk of the information situated around these centers. This strategy renders our proposed GB-RVFL model less susceptible to noise and outliers, which are typically situated farther away from the central data distribution or clusters. A comprehensive examination of the robustness of our proposed GB-RVFL model, in comparison to baseline models, has been conducted in Section \ref{subsec:noise}.
    \item \textbf{\textit{Scalability:}} By training the proposed GB-RVFL on GBs instead of the entire training dataset, and considering that the number of GBs is significantly smaller than the total training data points, we enhance the scalability of our model. For a deeper understanding and justification, please refer to Section \ref{subsec:complexity} for the complexity analysis of our proposed model and Section \ref{subsec:NDC} for the experiments on scalability investigation.
\end{itemize}

Let the hidden layer matrix is denoted by $\mathfrak{G}$, acquired through the random projection of the matrix of centers $\mathcal{O}$ of GBs followed by applying the activation function $\gamma$, defined as:
 \begin{align}
 \label{GB_G_1_layer}
        \mathfrak{G}=\gamma(\mathcal{O} \mathcal{X}  + \mathbf{1}^t \otimes \zeta) \in \mathbb{R}^{k \times g},
     \end{align}
where $\mathcal{X} \in \mathbb{R}^{P \times g} $ is the randomly initialized weights matrix and $\zeta$ $\in \mathbb{R}^{1 \times g}$ is the bias vector. 
The output layer's weights are calculated as follows:
\begin{equation}\label{eq:GB_train1}
[\mathfrak{G} \oplus \mathcal{O}]\Omega=D\Omega=\hat{\mathcal{W}}.   
\end{equation}
Here, $\Omega$ in $\mathbb{R}^{(P + g) \times C}$ represents the weights matrix that links the input layer and hidden layer (GBs) to the output layer (GBs). $\hat{\mathcal{W}}$ is the anticipated output and $D=[\mathfrak{G} \oplus \mathcal{O}] \in \mathbb{R}^{k \times (P+g)}$.

The proposed optimization problem for the proposed GB-RVFL model is formulated as follows: 
\begin{align}   
\label{eq20}
&\Omega_{\min}=\underset{{\Omega}}{\text{argmin}}\,  \frac{\mathcal{C}}{2}\|\phi\|_2^2+\frac{1}{2}\|\Omega\|_2^2, \nonumber \\
& \text{s.t.} ~~D \Omega-\mathcal{W}=\phi,
\end{align} 
where $\phi$ refers to the error matrix and $\mathcal{C}$ is the regularization parameter. 
Problem \eqref{eq20} is the convex quadratic programming problem (QPP) and hence possesses a unique solution. The Lagrangian of \eqref{eq20} is written as:
\begin{align}   
\label{eq22}
\mathcal{L}(\Omega,\phi,\lambda)=\frac{\mathcal{C}}{2}\|\phi\|_2^2+\frac{1}{2}\|\Omega\|_2^2-\lambda^t(D \Omega-\mathcal{W}-\phi),
\end{align}
where $\lambda$ is the Lagrangian multiplier. Differentiating $\mathcal{L}$ partially  w.r.t. each parameters, $i.e., \Omega,\phi ~\text{and}~ \lambda$; and equating them to zero, we obtain
\begin{align}   
& \frac{\partial \mathcal{L}}{\partial \Omega}=0 \Rightarrow \Omega - D^t \lambda = 0 \Rightarrow \Omega = D^t \lambda, \label{eq23}\\
& \frac{\partial \mathcal{L}}{\partial \phi}=0 \Rightarrow \mathcal{C}  \phi +\lambda = 0 \Rightarrow \lambda = - \mathcal{C} \phi,\label{eq24}\\
& \frac{\partial \mathcal{L}}{\partial \lambda}=0 \Rightarrow D \Omega-\mathcal{W}-\phi = 0 \Rightarrow \phi = D \Omega-\mathcal{W}. \label{eq25}
\end{align}
Substituting Eq. \eqref{eq25} in \eqref{eq24}, we get
\begin{align}  
\label{eq26}
\lambda = - \mathcal{C} (D \Omega-\mathcal{W}).
\end{align}
On substituting the value of $\lambda$ obtained in Eq. \eqref{eq23}, we obtain
\begin{align} 
&\Omega = D^t\left(- \mathcal{C} (D \Omega-\mathcal{W})\right), \label{eq27a} \\
 \Rightarrow ~ & \Omega = -\mathcal{C} D^t D \Omega + \mathcal{C} D^t \mathcal{W}, \label{eq27b}\\
 \Rightarrow ~ & \left(I + \mathcal{C} D^t D \right) \Omega = \mathcal{C} D^t \mathcal{W}, \label{eq27c}\\
 \Rightarrow ~ & \Omega = \left(D^tD + \frac{1}{\mathcal{C}} I\right)^{-1} D^t \mathcal{W}, \label{eq27d}
\end{align}
where $I$ is the identity matrix of the appropriate dimension. Substituting the values of \eqref{eq25} and \eqref{eq23} in \eqref{eq24}, we get
\begin{align} 
\label{eq29a}
&\lambda = -{\mathcal{C}} (DD^t\lambda -\mathcal{W}),\\
\Rightarrow ~ & \lambda + {\mathcal{C}} DD^t\lambda = \mathcal{C}\mathcal{W}, \label{eq29b}\\
\Rightarrow ~ & \mathcal{C}\left(\frac{1}{\mathcal{C}}I +  DD^t\right)\lambda = \mathcal{C}\mathcal{W}, \label{eq29c}\\
\Rightarrow ~ & \lambda = \left(\frac{1}{\mathcal{C}}I +  DD^t\right)^{-1}\mathcal{W}.\label{eq29d}
\end{align}
Putting the obtained value of  $\lambda$ from Eq. \eqref{eq29d} in \eqref{eq23}, we get
\begin{align} 
\Omega = D^t
\left(\frac{1}{\mathcal{C}}I+DD^t\right)^{-1}\mathcal{W}. \label{eq31}
\end{align}
We get two distinct formulas, \eqref{eq27d} and \eqref{eq31}, that can be utilized to determine $\Omega$. It is worth noting that both formulas involve the calculation of the matrix inverse. 
If the number of features ($P+g$) in $D$ is less than or equal to the number of GBs ($k$), we employ the formula \eqref{eq27d} to compute $\Omega$. Otherwise, we opt for the formula \eqref{eq31} to calculate $\Omega$. As a result, we possess the advantage of calculating the matrix inverse either in the feature or sample space, contingent upon the specific scenario. Therefore, the optimal solution of \eqref{eq20} is given as:
\begin{equation}
\label{eq:Omega_GB_RVFL}
    \Omega=\left\{\begin{array}{ll}{\left(D^tD + \frac{1}{\mathcal{C}} I \right)^{-1} D^t \mathcal{W}}, & \text{if}~~ (P+g) \leq k,  \vspace{3mm} \\ {D^t
\left(\frac{1}{\mathcal{C}} I+DD^t\right)^{-1}\mathcal{W}} , & \text{if}~~ k < (P+g).\end{array}\right. 
\end{equation}
\subsection{Graph Embedded Granular Ball Random Vector Functional Link (GE-GB-RVFL) Network}
This subsection delves into the proposed GE-GB-RVFL model, commencing with the establishment of its foundational mathematical framework. The GE-GB-RVFL model inherits all the properties of the proposed GB-RVFL model (i.e., robustness and scalability) along with preserving the inherent geometrical structure of GBs.
\begin{itemize}
    \item \textbf{\textit{Preservation of the original geometrical structure of granular balls:}} The preservation of the original geometrical structure of GBs is achieved through the optimization process (see Eq. \ref{GE-GB-eq:1}) for determining the GE-GB-RVFL network's output weights. This process incorporates subspace learning (SL) criteria, utilizing intrinsic and penalty SL within the GE and granular computation frameworks. The GE framework adeptly manages the geometric relationships among the centers of GBs. In contrast, the standard RVFL model lacks this capability, leading to a failure in preserving the original geometrical structure of datasets and, thus, losing valuable information during training.
\end{itemize}

The proposed optimization problem of the GE-GB-RVFL is articulated as follows:
\begin{align} 
\label{GE-GB-eq:1}
    \underset{\Omega}{\text{min}}~~ &\frac{1}{2}\|\Omega\|_2^2+\frac{\mathcal{C}}{2}\|\xi\|_2^2+\frac{\alpha}{2}\|U^{\frac{1}{2}}\Omega\|_2^2 \nonumber\\
    s.t.~~& D\Omega-\mathcal{W}=\xi. 
\end{align}
Here, $\Omega$ denotes the weights matrix that establishes connections between the input layer and hidden layer (GBs) and the output layer (GBs), $\xi$ is the error matrix, and $\mathcal{C}$ is the regularization parameter. Eq. \eqref{GE-GB-eq:1} assigns distinct significance to each center of the GBs by incorporating weights and considering the geometric relationships among GBs by including the GE term. Here, $U$ represents the GE matrix, and $\alpha$ denotes the graph regularization parameter. In this study, both the intrinsic and penalty graphs are defined over the concatenated matrix $D$. Following subsection \ref{subsection:G_matrix_Graph}, the intrinsic graph is denoted as $\mathcal{U}^{int}=\{D,\Theta^{int}\}$, and the penalty graph is represented as $\mathcal{U}^{pen}=\{D,\Theta^{pen}\}$. Consequently, $U_{int}=D^{t}\mathcal{L}D$, and ${U}_{pen}=D^{t}\mathcal{S}D$. Drawing from the literature, the weights for intrinsic and penalty graphs in the context of linear discriminant analysis (LDA) \cite{tharwat2017linear} are given as follows:
\begin{align}
        \label{LDAgraph1}
        \Theta^{int}_{ij} &= \left\{\begin{array}{ll} \frac{1}{G_{w_i}}, & \text{if}~~ w_i = w_j ,\vspace{3mm} \\ 
0, & \text{otherwise}. \end{array}\right.\\
        \label{LDAgraph11}
        \Theta^{pen}_{ij} &= \left\{\begin{array}{ll} \frac{1}{k}-\frac{1}{G_{w_i}}, & \text{if}~~ w_i = w_j ,\vspace{3mm} \\ 
\frac{1}{k}, & \text{otherwise}. \end{array}\right.
    \end{align}
    Here, $G_{w_i}$ denotes the number of GBs (centers) with class label $w_i$.\\
The Lagrangian of \eqref{GE-GB-eq:1} is written as: 
\begin{align} 
\label{GE-GB-eq:2}
    \mathcal{L}_{GE}=\frac{1}{2}\|\Omega\|_2^2+\frac{\mathcal{C}}{2}\|D\Omega-\mathcal{W}\|_2^2+\frac{\alpha}{2}\|U^{\frac{1}{2}}\Omega\|_2^2.
\end{align}
Applying the Karush-Kuhn-Tucker (K.K.T.) condition in \eqref{GE-GB-eq:2}, we get following:
\begin{align}  
\frac{\partial {\mathcal{L}_{GE}}}{\partial \Omega}= \mathcal{C}D^{t}(D\Omega-\mathcal{W})+\Omega+\alpha{U}\Omega = 0.
\end{align}
Upon computation, the output layer parameters are obtained as follows:
\begin{align}  
\label{eq:GE-GB-RVFL_weights}
\Omega= \left(D^{t}D + \frac{1}{\mathcal{C}}I + \frac{\alpha}{\mathcal{C}}U\right)^{-1}D^{t}\mathcal{W}.
\end{align}
The matrix $U$ can be defined in two cases:
\begin{enumerate}
    \item Case 1: When $U$ represents intrinsic training data relationships ($U_{pen} = I$), it indicates that no penalties are imposed on relationships among the nodes (GB centers). This approach solely relies on intrinsic relationships to extract graphical information from the GBs.
    \item Case 2: When $U$ encompasses both intrinsic and penalty training data relationships ($U_{pen} \neq I$), specific penalties are applied to certain relationships among the nodes (GB centers). Here, the emphasis is on utilizing both intrinsic and penalty-specific relationships to extract graphical information from the GBs.
\end{enumerate}
\textbf{\textit{Remarks}} 
\begin{enumerate}
    \item In our proposed GE-GB-RVFL model, the embedding space of the Graph $U$ is in the GB-RVFL space $\mathbb{R}^{(P+g)}$, rather than the input space $\mathbb{R}^P$, provides a more accurate representation of both linear and nonlinear relationships among the nodes/centers of GBs.
    \item Incorporating graph embedding (GE) increases the complexity of the proposed GE-GB-RVFL model. Therefore, future research could explore alternative methods, such as sparse GE techniques, to preserve the geometrical structure of datasets while reducing computational overhead.
    \item In our experiments, we used LDA for the graph embedding, as it is particularly effective for maximizing class separability in supervised learning tasks. However, other techniques such as, local Fisher discriminant analysis (LFDA) or marginal Fisher analysis (MFA) \cite{yan2006graph}, can also be explored, depending on the task requirements.
\end{enumerate}
The algorithms of the proposed GB-RVFL and GE-GB-RVFL models are given in Algorithm \ref{Algorithm_GB_RVFL.}.
\begin{algorithm}[ht!]
\caption{Algorithm of the proposed GB-RVFL and GE-GB-RVFL models.}
\label{Algorithm_GB_RVFL.}
\textbf{Input:} Traning dataset $T$, and the threshold purity $\rho$. \\
\textbf{Output:} Output weights of the GB-RVFL.\\ \vspace{-5mm}
\begin{algorithmic}[1]
\STATE Assume the entire dataset $T$ as a granular ball $GB$ and set of GBs, $G$, to be empty set, i.e., $GB=T$ and $G=\{\hspace{0.1cm}\}$. 
\STATE $Temp =\{GB\}$.
\STATE $for$ $i = 1:\lvert Temp \rvert$ 
\STATE $if$ $pur(GB_i) < \rho$ 
\STATE Split $GB_i$ into $GB_{i1}$ and $GB_{i2}$, using $2$-means clustering algorithm. 
\STATE $Temp \leftarrow GB_{i1}, \hspace{0.05cm} GB_{i2}$. 
\STATE $end$ $if$. 
\STATE $else$ $pur(GB_i)\geq \rho$ 
\STATE Calculate the center $o_i = \frac{1}{p} \sum_{j=1}^{p} v_j$ of $GB_i$, where $v_j \in GB_i$, $j=1, 2, \ldots, p$, and $p$ is the number of training sample in $GB_i$.  
\STATE Calculate the label $w_i$ of $GB_i$, where $w_i$ is assigned the label of majority class samples within $GB_i$.
\STATE Put $GB_i = \{(o_i,w_i)\}$ in $G$. 
\STATE $end$ $else.$ 
\STATE $end$ $for.$  
\STATE $if$ $Temp \neq \{\hspace{0.1cm}\}$ 
\STATE Go to step 3 (for further splitting). 
\STATE $end$ $if.$ 
\STATE Set $G=\{GB_i, \hspace{0.2cm} i=1,2, \ldots, k\} = \{(o_i, w_i),\hspace{0.2cm} i=1,2, \ldots, k\},$ where $o_i$ signifies the center of the GB, $w_i$ is the label of $GB_i$ and $k$ is the number of generated GBs. 
\STATE Find the hidden layer features using \eqref{GB_G_1_layer}.
\STATE Create the enhanced features using \eqref{eq:GB_train1}. 
\STATE \textbf{For the output weights of GB-RVFL model:} Calculate the output layer weights using \eqref{eq:Omega_GB_RVFL}.
\STATE \textbf{For the output weights of GE-GB-RVFL model:} Calculate intrinsic and penalty graphs using \eqref{LDAgraph1} and \eqref{LDAgraph11}, respectively, then the output layer weights are calculated using \eqref{eq:GE-GB-RVFL_weights}.
\end{algorithmic}
\end{algorithm}
\subsection{Time and Space Complexity Analysis of the Proposed Models}
\label{subsec:complexity}
Here, we discuss the time and space complexity of the proposed GB-RVFL and GE-GB-RVFL models.
\textbf{Time Complexity of GB-RVFL:} The complexity of the proposed GB-RVFL model primarily depends on (a) the requirement of matrix inversion in \eqref{eq:Omega_GB_RVFL}, and (b) GB computation. Following \cite{shi2021random}, time complexity in computing inverse in RVFL is $\mathcal{O}(M^3)$ if $M \leq (P+g)$ or $\mathcal{O}((P+g)^3)$ if $(P+g) < M$. 
Therefore, the time complexity to find the inverse in \eqref{eq:Omega_GB_RVFL} is  $\mathcal{O}(k^3)$ if $k \leq (P+g)$ or $\mathcal{O}((P+g)^3)$ if $(P+g) < k$. In the generation of GB, we use 2-means clustering; therefore, in 2-means clustering, the time complexity is $\mathcal{O}(2M(iter))$ \cite{zhou2009novel}, where $M$ represents the number of samples in the training dataset $T$, and $iter$ is the number of iterations. Our approach starts with the training dataset $T$, which we treat as the initial granular ball ($GB$) set. Using the $2$-means clustering method, we divide $GB$ into two GBs initially, with a computational complexity of $\mathcal{O}(2M)$. In subsequent phases, if both GBs are impure, they undergo further division into four GBs, maintaining a maximum computational complexity of $\mathcal{O}(2M)$ each time. This iterative process continues for a total of $iter$ iterations. The overall computational complexity of generating GBs is $\mathcal{O}(2M(iter))$ or less, considering the maximum computational complexity per iteration and the total number of iterations $iter$. Thus, the time complexity of the proposed GB-RVFL is $\mathcal{O}(k^3)$+$\mathcal{O}(2M(iter))$ if $k \leq (P+g)$ or $\mathcal{O}((P+g)^3)$+$\mathcal{O}(2M(iter))$ if $(P+g) < k$. Since, $k<<M$, therefore, $\mathcal{O}(k^3)<<\mathcal{O}(M^3)$. This implies that if the inverse of the matrix is calculated in the sample space, then the time complexity of the proposed GB-RVFL is much less than the RVFL model. \\
\textbf{Time Complexity of GE-GB-RVFL:} Additionally, the time complexity of the proposed GE-GB-RVFL depends upon one more factor, which is the computation of the GE matrix $U$. This matrix accounts for both intrinsic and penalty graph structures. According to \cite{ganaie2024graph}, the time complexity for this computation is $\mathcal{O}((P + g)^3 + (P + g)^2k)$. Thus, the time complexity of the proposed GB-RVFL is $\mathcal{O}(k^3)$+$\mathcal{O}(2M(iter))$+$\mathcal{O}((P + g)^3 + (P + g)^2k)$ if $k \leq (P+g)$ or $\mathcal{O}((P+g)^3)$+$\mathcal{O}(2M(iter))$+$\mathcal{O}((P + g)^3 + (P + g)^2k)$ if $(P+g) < k$.\\
\textbf{Space complexity of GB-RVFL:} The space complexity of the GB-RVFL model can be understood by considering the storage requirements for its components. First, the model requires space to store the centers of the GBs, which involves $k$ GBs, each with $P$ features, resulting in a complexity of $\mathcal{O}(k P)$. Additionally, the hidden layer weights matrix, involving $g$ hidden nodes and $P$ input features, contributes $\mathcal{O}(P  g)$. Furthermore, the inversion of the GB center matrix, which is of size $k \times g$, requires $\mathcal{O}(k g)$. Therefore, the overall space complexity of the GB-RVFL model is $\mathcal{O}(kP + Pg)$.\\
\textbf{Space complexity of GE-GB-RVFL:} For the proposed GE-GB-RVFL model, which extends GB-RVFL by incorporating graph embedding (GE), additional storage is required. The graph embedding involves an adjacency matrix to capture relationships between data points of $P+g$ features of the matrix $D$, leading to a potential space complexity of $\mathcal{O}((P+g)^2)$. Consequently, the total space complexity of the GE-GB-RVFL model is $\mathcal{O}(kP + Pg)$ + $\mathcal{O}((P+g)^2)$.
\section{Numerical Experiments and Results}
\label{experiments}
This section presents comprehensive details of the experimental setup, datasets, and compared models. Subsequently, we delve into the experimental results and conduct statistical analyses. We also examine the influence of noise on the performance of the proposed GB-RVFL and GE-GB-RVFL models.
\subsection{Datasets}
To evaluate the efficacy of the proposed GB-RVFL and GE-GB-RVFL models, we employ 30 benchmark datasets from UCI \cite{dua2017uci} and KEEL \cite{derrac2015keel}. To test the scalability of the proposed model, we use NDC \cite{ndcdata1998} dataset, generated using David Musicant's NDC Data Generator, encompasses samples ranging from 50 thousand to 100 million, consistently featuring 32 features. Moreover, as an application, we employ our proposed models to classify the various stages of Alzheimer's disease using the ADNI\footnote{\url{https://adni.loni.usc.edu/}}dataset and breast cancer detection using the BreakHis dataset \cite{spanhol2015dataset}.
\subsection{Compared Models}
We conduct comparisons among the proposed GB-RVFL and GE-GB-RVFL; and several benchmarks, including the standard RVFL \cite{pao1994learning}, RVFL without direct link (RVFLwoDL) (also known as Extreme Learning Machine (ELM)) \cite{huang2006extreme}, Intuitionistic Fuzzy RVFL (IF-RVFL) \cite{malik2022alzheimer}, Neuro-fuzzy RVFL (NF-RVFL) \cite{sajid2024neuro}, and RVFL based on Wave loss function (Wave-RVFL) \cite{sajid2024wavervfl}. Further, we formulate and propose GB-RVFLwoDL (granular ball RVFL without direct link) to test the significance of direct links. The formulation of GB-RVFLwoDL is the same as GB-RVFL, except the direct links connection is missing in GB-RVFLwoDL.
\subsection{Experimental Setup and Hyperparameter Setting}
The experimental configuration includes a PC equipped with an Intel(R) Xeon(R) Gold 6226R CPU running at a speed of $2.90$GHz and featuring $128$ GB of RAM. This system runs on the Windows 11 platform and executes tasks using Python $3.11$. The dataset is randomly divided into training and testing subsets at a ratio of $70:30$, respectively. We utilize a five-fold cross-validation technique combined with a grid search approach to optimize the hyperparameters of the models within specified ranges: $\mathcal{C}=\alpha = \{10^{-5}, 10^{-4}, \ldots, 10^{5}\}$. The number of hidden nodes is selected within the range $3:20:203$. Following \cite{sajid2024neuro}, for the NF-RVFL model, the neuro-fuzzy layer's fuzzy rules are chosen from the range $J=5 : 5 : 50$ and $k$-means clustering is used to generate centres of the neuro-fuzzy layer. $\mu$ for the membership function in the IFRVFL model is chosen from the set $\{2^{-5},2^{-3},\hdots,2^5\}$. 
The wave loss parameters used for the Wave-RVFL model are the same as those in \cite{sajid2024wavervfl}.
For all models, we tune $10$ activation functions with indices ranging from $1$ to $10$. The details are provided in Table S.1 of the supplementary material.
\subsection{Results and Discussions on UCI and KEEL Datasets}
The proposed GB-RVFL and GE-GB-RVFL models are evaluated against standard RVFL, RVFLwoDL, GB-RVFLwoDL, IF-RVFL, Wave-RVFL and NF-RVFL models. The accuracy and rank of the models are reported in Table \ref{tab:uci_and_keel}. The best hyperparameter settings for all the compared models on UCI and KEEL datasets are reported in Table S.2 of the supplementary material. We compare the performance of the proposed models using accuracy and statistical tests (following recommendations of \citet{demvsar2006statistical}), including the ranking, Friedman, and Nemenyi post hoc tests.
\subsubsection{Accuracy} 
As per Table \ref{tab:uci_and_keel}, the proposed GE-GB-RVFL achieved the highest average accuracy at $85.03\%$, with the proposed GB-RVFL following closely at $84.44\%$. The number of fuzzy-based models, i.e., IF-RVFL and NF-RVFL, as well as RVFL, RVFLwoDL, Wave-RVFL, and GB-RVFLwoDL models, comes later than the proposed models. This finding suggests that the proposed GB-RVFL and GE-GB-RVFL models mitigate the detrimental effect of noise and outliers and simultaneously leverage the original topological structure of the center of the GBs.
\begin{table*}[ht!]
\centering
\caption{Accuracy and rank of the proposed GB-RVFL and GE-GB-RVFL models against the baseline models on real-world datasets, i.e., KEEL and UCI.}
\label{tab:uci_and_keel}
\resizebox{\textwidth}{!}{%
\begin{tabular}{l|cccccccc}\hline \vspace{-2mm}\\ 
 \text{Model} $\rightarrow$ & RVFL \cite{pao1994learning}                 & RVFLwoDL  \cite{huang2006extreme}  & IF-RVFL \cite{malik2022alzheimer}
& NF-RVFL \cite{sajid2024neuro} & Wave-RVFL \cite{sajid2024wavervfl}      & GB-RVFLwoDL$^{\bigstar}$                 & GB-RVFL$^{\bigstar}$                & GE-GB-RVFL$^{\bigstar}$   
                           \vspace{1mm}\\ \hline \vspace{-2mm}\\
            Dataset $\downarrow$ &
  \multicolumn{1}{l}{(ACC, Rank)} &
  \multicolumn{1}{l}{(ACC, Rank)} &
  \multicolumn{1}{l}{(ACC, Rank)} &
  \multicolumn{1}{l}{(ACC, Rank)} &
  \multicolumn{1}{l}{(ACC, Rank)} &
  \multicolumn{1}{l}{(ACC, Rank)} &
  \multicolumn{1}{l}{(ACC, Rank)} &
  \multicolumn{1}{l}{(ACC, Rank)}  \vspace{1mm}\\ \hline \vspace{-2mm}\\
acute\_inflammation &
  ($100, 3$) &
  ($100, 3$) &
  ($83.33, 6$) &
  ($75.56, 7$) &
  ($100, 3$) &
  ($75, 8$) &
  ($100, 3$) &
  ($100, 3$) \\
aus &
  ($86.94, 6$) &
  ($87.98, 1.5$) &
  ($86.54, 8$) &
  ($87.98, 3$) &
  ($87.98, 1.5$) &
  ($87.02, 4.5$) &
  ($87.02, 4.5$) &
  ($86.54, 7$) \\
bank &
  ($89.17, 3$) &
  ($89.31, 2$) &
  ($88.14, 8$) &
  ($88.95, 4$) &
  ($88.52, 6$) &
  ($88.87, 5$) &
  ($88.43, 7$) &
  ($89.83, 1$) \\
chess\_krvkp &
  ($90.41, 4$) &
  ($90.2, 5$) &
  ($90.62, 3$) &
  ($80.95, 8$) &
  ($87, 7$) &
  ($88.95, 6$) &
  ($91.45, 2$) &
  ($92.49, 1$) \\
cleve &
  ($81.11, 3$) &
  ($80, 4$) &
  ($78.89, 5$) &
  ($82.22, 1$) &
  ($75.46, 8$) &
  ($76.67, 6$) &
  ($75.56, 7$) &
  ($82.22, 2$) \\
conn\_bench\_sonar\_mines\_rocks &
  ($74.6, 3.5$) &
  ($71.43, 7$) &
  ($70.95, 8$) &
  ($74.6, 6$) &
  ($74.63, 1$) &
  ($74.6, 3.5$) &
  ($74.6, 3.5$) &
  ($74.6, 3.5$) \\
crossplane130 &
  ($97.44, 6.5$) &
  ($97.44, 6.5$) &
  ($94.87, 8$) &
  ($100, 3$) &
  ($100, 3$) &
  ($100, 3$) &
  ($100, 3$) &
  ($100, 3$) \\
echocardiogram &
  ($85, 6$) &
  ($87.5, 2.5$) &
  ($81, 7$) &
  ($80, 8$) &
  ($86.5, 5$) &
  ($87.5, 2.5$) &
  ($87.5, 2.5$) &
  ($87.5, 2.5$) \\
ecoli-0-1-4-6\_vs\_5 &
  ($98.81, 4.5$) &
  ($98.81, 4.5$) &
  ($98.81, 2$) &
  ($94.05, 8$) &
  ($97.93, 7$) &
  ($100, 1$) &
  ($98.81, 4.5$) &
  ($98.81, 4.5$) \\
ecoli0137vs26 &
  ($85.74, 7$) &
  ($86.81, 5$) &
  ($86.62, 6$) &
  ($88.68, 2$) &
  ($87.18, 4$) &
  ($58.51, 8$) &
  ($87.23, 3$) &
  ($89.36, 1$) \\
ecoli2 &
  ($82.08, 7$) &
  ($81.09, 8$) &
  ($85.12, 5.5$) &
  ($85.12, 5.5$) &
  ($86.38, 3$) &
  ($85.15, 4$) &
  ($87.13, 1.5$) &
  ($87.13, 1.5$) \\
fertility &
  ($90, 2$) &
  ($80, 7$) &
  ($83.33, 6$) &
  ($73.33, 8$) &
  ($86.33, 5$) &
  ($90, 2$) &
  ($86.67, 4$) &
  ($90, 2$) \\
haberman\_survival &
  ($76.09, 5$) &
  ($78.26, 3$) &
  ($73.91, 7$) &
  ($71.74, 8$) &
  ($74.47, 6$) &
  ($78.26, 3$) &
  ($78.26, 3$) &
  ($79.35, 1$) \\
heart-stat &
  ($81.89, 3.5$) &
  ($81.89, 3.5$) &
  ($79.78, 5.5$) &
  ($79.78, 5.5$) &
  ($72.19, 8$) &
  ($77.78, 7$) &
  ($82.72, 1$) &
  ($82.72, 2$) \\
heart\_hungarian &
  ($72.78, 7$) &
  ($72.65, 8$) &
  ($77.53, 3$) &
  ($87.65, 2$) &
  ($96.61, 1$) &
  ($73.03, 6$) &
  ($74.16, 5$) &
  ($75.28, 4$) \\
ionosphere &
  ($82.79, 5$) &
  ($82.45, 6$) &
  ($87.74, 1.5$) &
  ($87.74, 1.5$) &
  ($74, 8$) &
  ($75.47, 7$) &
  ($83.96, 3$) &
  ($83.02, 4$) \\
led7digit-0-2-4-5-6-7-8-9\_vs\_1 &
  ($94.74, 3.5$) &
  ($94.74, 3.5$) &
  ($86.47, 8$) &
  ($92.48, 7$) &
  ($92.85, 6$) &
  ($95.49, 1$) &
  ($93.99, 5$) &
  ($94.99, 2$) \\
mammographic &
  ($82.01, 4$) &
  ($82.35, 3$) &
  ($81.66, 6$) &
  ($80.28, 7.5$) &
  ($81.97, 5$) &
  ($80.28, 7.5$) &
  ($83.74, 1$) &
  ($83.39, 2$) \\
monk1 &
  ($42.52, 6$) &
  ($42.1, 7$) &
  ($49.1, 1$) &
  ($44.31, 5$) &
  ($44.31, 3.5$) &
  ($41.32, 8$) &
  ($44.31, 3.5$) &
  ($47.9, 2$) \\
monks\_3 &
  ($90.41, 2.5$) &
  ($90.41, 2.5$) &
  ($87.9, 4$) &
  ($85.63, 6.5$) &
  ($85.63, 6.5$) &
  ($85.63, 6.5$) &
  ($85.63, 6.5$) &
  ($91.02, 1$) \\
new-thyroid1 &
  ($90, 5.5$) &
  ($90, 5.5$) &
  ($98.46, 2$) &
  ($100, 1$) &
  ($89.44, 7$) &
  ($70.77, 8$) &
  ($90.77, 4$) &
  ($93.85, 3$) \\
pima &
  ($74.03, 5$) &
  ($73.59, 6$) &
  ($75.76, 4$) &
  ($76.19, 3$) &
  ($70.8, 8$) &
  ($71.86, 7$) &
  ($79.7, 1$) &
  ($76.62, 2$) \\
statlog\_heart &
  ($80.89, 5.5$) &
  ($80.89, 5.5$) &
  ($81.36, 4$) &
  ($87.65, 1$) &
  ($80.37, 7$) &
  ($77.78, 8$) &
  ($82.72, 2$) &
  ($82.72, 3$) \\
tic\_tac\_toe &
  ($96.65, 7$) &
  ($96.65, 7$) &
  ($99.65, 1.5$) &
  ($99.65, 1.5$) &
  ($96.65, 7$) &
  ($97.22, 5$) &
  ($97.57, 4$) &
  ($99.31, 3$) \\
transfusion &
  ($76.44, 2$) &
  ($71.11, 8$) &
  ($74.56, 7$) &
  ($79.56, 1$) &
  ($74.89, 6$) &
  ($75.11, 5$) &
  ($76, 3.5$) &
  ($76, 3.5$) \\
vehicle2 &
  ($90.85, 6$) &
  ($90.03, 8$) &
  ($97.64, 1.5$) &
  ($97.64, 1.5$) &
  ($90.64, 7$) &
  ($91.34, 5$) &
  ($91.73, 4$) &
  ($92.52, 3$) \\
vertebral\_column\_2clases &
  ($81.4, 7$) &
  ($82.25, 6$) &
  ($87.1, 3$) &
  ($84.95, 4$) &
  ($94.84, 1$) &
  ($82.37, 5$) &
  ($74.19, 8$) &
  ($90.32, 2$) \\
wpbc &
  ($69.49, 5.5$) &
  ($70.97, 4$) &
  ($76.27, 3$) &
  ($79.66, 1$) &
  ($68.36, 7$) &
  ($62.71, 8$) &
  ($76.27, 2$) &
  ($69.49, 5.5$) \\
yeast-0-2-5-7-9\_vs\_3-6-8 &
  ($95.35, 6$) &
  ($95.68, 5$) &
  ($96.03, 2$) &
  ($95.7, 4$) &
  ($95.9, 3$) &
  ($94.37, 7$) &
  ($98.01, 1$) &
  ($77.81, 8$) \\
yeast1 &
  ($74.01, 8$) &
  ($74.78, 6.5$) &
  ($75.34, 3$) &
  ($74.78, 6.5$) &
  ($75.09, 5$) &
  ($75.78, 2$) &
  ($75.11, 4$) &
  ($76.23, 1$) \\ \hline
Average (ACC, Rank) &
  ($83.79, 4.98$) &
  ($83.38, 5.13$) &
  ($83.82, 4.65$) &
  ($83.89, 4.37$) &
  ($83.9, 5.18$) &
  ($80.63, 5.32$) &
  ($84.44, 3.57$) &
  ($\textbf{85.03}, \textbf{2.8}$) \vspace{1mm}\\ \hline
\multicolumn{9}{l}{The best-performing model in terms of average accuracy is indicated by boldface in the last row. ACC is an acronym for accuracy. $^{\bigstar}$ denotes the proposed models.}\\
\end{tabular}%
}
\end{table*}
\begin{table}[ht!]
\centering
\caption{Comparison of the proposed GE-GB-RVFL and other compared models using Nemenyi post hoc test.}
\label{tab:Nemenyi_post_hoc_test}
\resizebox{15cm}{!}{
\resizebox{\textwidth}{!}{%
\begin{tabular}{lcccccc} \hline
                & RVFL \cite{pao1994learning} & RVFLwoDL \cite{huang2006extreme}  & IF-RVFL \cite{malik2022alzheimer} & NF-RVFL \cite{sajid2024neuro} & Wave-RVFL \cite{sajid2024wavervfl} &GB-RVFLwoDL$^{\bigstar}$ \\ \hline
Rank Difference & 2.18 & 2.33 & 1.85    & 1.57 & 2.38   & 2.52          \\ \hline
Significance    & Yes  & Yes  & Yes     & No   & Yes   & Yes  
\vspace{1mm}\\ \hline 
\multicolumn{7}{l}{The first row presents the difference in ranking between the proposed GE-GB-RVFL model and the models listed in}\\
\multicolumn{7}{l}{ the respective columns, whereas the second row shows the significant superiority of the proposed GE-GB-RVFL model }\\
\multicolumn{7}{l}{listed in the respective columns. $^{\bigstar}$ denotes the proposed model.}\\
\end{tabular}}}
\end{table}
\subsubsection{Statistical rank}In the ranking methodology, models are ranked according to their performance across distinct datasets. Poorer performers receive higher ranks, while top performers receive lower ranks. Let's envision a situation with $\mathfrak{M}$ models evaluated across $\mathfrak{D}$ datasets, where the rank of the $m$-th model on the $d$-th dataset is denoted as $\mathfrak{R}_m^d$. Mathematically, the average rank of the $m$-th model can be calculated as: $\mathfrak{R}_m=\left(\sum_{d=1}^{\mathfrak{D}}\mathfrak{R}_m^d\right)/\mathfrak{D}$. Each model's rank on every dataset are presented in Table \ref{tab:uci_and_keel}. The results from Table \ref{tab:uci_and_keel} highlight a consistent trend that both the proposed GB-RVFL and GE-GB-RVFL models consistently secure the lowest ranks in most of the datasets. Further, the GE-GB-RVFL and GB-RVFL's average ranks are $2.8$ and $3.57$, respectively. These ranks position our proposed models as the top two lowest-ranking models among those compared. This emphasizes the superior generalization performance of the proposed GB-RVFL and GE-GB-RVFL models. In contrast, GB-RVFLwoDL performs least favorably, with an average rank of $5.32$. This underscores the clear superiority of the proposed GB-RVFL and GE-GB-RVFL models over the GB-RVFLwoDL and baseline models, which further shows the significance of direct links in the performance of the RVFL based-models.
\subsubsection{Friedman test} To gain deeper statistical insights and compare the average ranks of models while identifying significant differences in their rankings, we employ the Friedman test \cite{friedman1937use}. Using a chi-squared statistic ($\chi^2_F$) with $(\mathfrak{M}-1)$ degrees of freedoms (DoFs), the test is formulated as follows: $\chi^2_F = \frac{12\mathfrak{D}}{\mathfrak{M} (\mathfrak{M}+1)} \left(\sum_{m=1}^{\mathfrak{M}} \left(\mathfrak{R}_m\right)^2 - \frac{\mathfrak{M}(\mathfrak{M}+1)^2}{4}\right)$ and the $F_F$ statistic is determined as:
$F_F=\chi_F^2\left(\frac{(\mathfrak{D}-1)}{\mathfrak{D}(\mathfrak{M}-1)-\chi_F^2}\right).$ The $F_F$ statistic's distribution has $(\mathfrak{M}-1)$ and $(\mathfrak{D}-1)(\mathfrak{M}-1)$ DoFs. In our experiment, $\mathfrak{M}=8$ and $\mathfrak{D}=30$, therefore, we get $\chi^2_F=27.78$ and $F_F=4.42$. Referring to the $F$-distribution table, we find that $F_F (7, 203) = 2.05$ at a $5\%$ significance level. Given that the calculated value of $4.42$ exceeds $2.05$, we reject the null hypothesis, indicating substantial differences among the models.
\subsubsection{Nemenyi post hoc test} Furthermore, we utilized the Nemenyi post hoc test \cite{demvsar2006statistical} to evaluate the significance of disparities between pairs of models. The critical difference ($C.D.$) is computed using the formula $C.D. = q_\alpha\left(\sqrt{\frac{\mathfrak{M}(\mathfrak{M} + 1)}{6\mathfrak{D}}}\right)$, with $q_\alpha$ indicating the critical value for the two-tailed Nemenyi test. With a calculated value of $C.D.=1.76$ (at $0.1$ level of significance ), models are considered significantly different if their average ranks differ by a $C.D.$ or greater. We compare our best-performed model with the other models, and corresponding results are reported in Table \ref{tab:Nemenyi_post_hoc_test}.
The rank differences between the proposed GE-GB-RVFL model and the RVFL, RVFLwoDL, IF-RVFL, Wave-RVFL, and GB-RVFLwoDL models are $2.18$, $2.33$, $1.85$, $2.38$ and $2.52$ respectively, all of which exceed the $C.D.$ value. This significant variance strongly reinforces the conclusion that the GE-GB-RVFL model outperforms the baseline and GB-RVFLwoDL models in a statistically significant manner. While the GE-GB-RVFL model does not demonstrate statistical superiority over the NF-RVFL model, its consistently lower average rank compared to the NF-RVFL model substantiates its overall superiority. Therefore, the utilization of GBs as inputs in the proposed model plays a very crucial role in the enhancement of the generalization performance of the proposed GB-RVFL and GE-RB-RVFL models over the baseline and GB-RVFLwoDL models.
\subsubsection{Direct link significance} Notably, the rank difference between GB-RVFL and GB-RVFLwoDL is $1.75$; this affirms the significance of direct links in the model's generalization performance. 
\\

\noindent \textbf{Main takeaways from the above discussions:}
\begin{itemize}
    \item The proposed GB-RVFL and GE-GB-RVFL models excel in both average accuracy and ranks.
    \item The superior accuracy and rank of GB-RVFL over GB-RVFLwoDL indicates that the direct links significantly contribute to the model's performance.
    \item Statistical analyses, including the Friedman and Nemenyi post hoc test, validate the statistical superiority of the proposed models compared to the existing baseline models.
\end{itemize}
\subsection{Results and Discussions on UCI and KEEL Datasets in Noisy Environment}
\label{subsec:noise}
We contaminate the label noise in the UCI and KEEL datasets at values of $5\%$, $10\%$, $20\%$, $30\%$, and $40\%$ to disturb the label of the samples in order to evaluate how robust the proposed GB-RVFL and GE-GB-RVFL models are against noise. In total, we take 7 datasets for this experiment. 
\subsubsection{Comparision among proposed, RVFL and RVFLwoDL models}
First of all, we compare our proposed models with baseline models, i.e., RVFL and RVFLwoDL  using $4$ datasets to test the robustness of the models. Among the $4$ datasets, two datasets (ecoli-0-1-4-6\_vs\_5 and yeast-0-2-5-7-9\_vs\_3-6-8) are chosen in such a manner that the proposed GE-GB-RVFL model does not top at $0\%$ noise. In the other two datasets, i.e., statlog\_heart and conn\_bench\_sonar\_mines\_rocks,  the proposed GE-GB-RVFL model tops at $0\%$ noise.

\begin{table}[ht!]
\centering
\caption{Classification accuracies of the proposed GB-RVFL and GE-GB-RVFL models against the baseline models on UCI and KEEL datasets with contaminated label noise.}
\label{tab:Noise}
\resizebox{\textwidth}{!}{
\begin{tabular}{l|c|cccccc} \hline \vspace{-2mm}\\ 
  \text{Model} $\rightarrow$      &      & RVFL \cite{pao1994learning}    & RVFLwoDL \cite{huang2006extreme} & Wave-RVFL \cite{sajid2024wavervfl}    & GB-RVFLwoDL$^{\bigstar}$  & GB-RVFL$^{\bigstar}$ & GE-GB-RVFL$^{\bigstar}$ \vspace{1mm}\\ \hline \vspace{-2mm}\\ 
    \text{Dataset} $\downarrow$    &  Noise  & ACC   & ACC    & ACC     & ACC  & ACC & ACC \vspace{1mm}\\ \hline \vspace{-2mm}\\
{conn\_bench\_sonar\_mines\_rocks} & 5\% & 76.1905 & 63.4921 & 70.63 & 71.4286 & 71.4286 & 76.1905 \\
        & 10\% & 70.1905 & 70.7778 & 70.88 & 57.1429 & 69.8413 & 73.0159    \\
        & 20\% & 70.9524 & 76.1905 & 77.63 & 71.4286 & 74.6032 & 77.7778    \\
        & 30\% & 71.4286 & 73.0159 & 71.88 & 82.5397 & 71.4286 & 76.1905    \\
        & 40\% & 66.6667 & 52.381  & 67  & 68.254  & 69.8413 & 69.8413   \\
Average &      & 71.0857 & 67.1714  & 71.604 & 70.1587 & \underline{71.4286} & \textbf{74.6032}    \vspace{1mm}\\ \hline \vspace{-2mm}\\
{ecoli-0-1-4-6\_vs\_5}               & 5\% & 90      & 90  & 92.5    & 94.0476 & 83.3333 & 94.0476 \\
        & 10\% & 98.8095 & 98.8095 & 97.88 & 100     & 98.8095 & 94.0476    \\
        & 20\% & 98.8095 & 98.8095  & 98.88 & 98.8095 & 98.8095 & 100        \\
        & 30\% & 90.8571 & 90.0476  &  91.88 & 75      & 94.0476 & 96.4286    \\
        & 40\% & 64.0476 & 61.6667  &  63.5 & 57.1429 & 65.4762 & 58.3333    \\
Average &      & 88.5048 & 87.8667  &  \underline{88.928} & 85      & 88.0952 & \textbf{88.5714}     \vspace{1mm}\\ \hline \vspace{-2mm}\\
{statlog\_heart}                   & 5\% & 88.8889 & 90.1235  & 81.88 & 81.4815 & 84.321  & 81.4815 \\
        & 10\% & 85.1235 & 85.1235 & 85.75 & 69.1358 & 86.4198 & 86.4198    \\
        & 20\% & 82.716  & 85.1852 & 85.75 & 83.9506 & 82.716  & 76.5432    \\
        & 30\% & 81.9506 & 81.9506 & 82.5 & 81.4815 & 82.716  & 99.6528    \\
        & 40\% & 56.7901 & 50.6173 & 59.75 & 62.963  & 77.7778 & 66.6667    \\
Average &      & 79.0938 & 78.6 & 79.126   & 75.8025 & \textbf{82.7901} & \underline{82.1528}    \vspace{1mm}\\ \hline \vspace{-2mm}\\
{yeast-0-2-5-7-9\_vs\_3-6-8}         & 5\% & 97.6821 & 96.6887 & 96.89 & 90.0662 & 97.6821 & 89.7351 \\
        & 10\% & 95.0132 & 96.6887 & 96.78 & 88.4106 & 97.0199 & 94.3709    \\
        & 20\% & 92.3576 & 92.0265 & 92.48 & 96.0265 & 95.3642 & 95.0331    \\
        & 30\% & 90.6887 & 90.6887  & 91.84 & 91.3907 & 92.7152 & 94.3709    \\
        & 40\% & 71.7219 & 71.7219  & 72.68 & 74.1722 & 73.8411 & 73.1788    \\
Average &      & 89.4927 & 89.5629 & \underline{90.134} & 88.0132 & \textbf{91.3245} & 89.3377     \vspace{1mm}\\ \hline \vspace{-2mm}\\
Overall Average ACC          &     & 84.015  & 82.7865 & 82.848 & 81.5653 & \textbf{85.7265} & \underline{85.2408}  \vspace{1mm}\\ \hline 
\multicolumn{8}{l}{
The best-performing model in terms of accuracy is indicated by boldface in each row.  ACC is an acronym for accuracy. The best hyperparameters}\\
\multicolumn{8}{l}{are reported in Table S.3 of the supplementary material. $^{\bigstar}$ denotes the proposed models.}\\
\end{tabular}}
\end{table}
\begin{figure*}[ht!]
\begin{minipage}{.5\linewidth}
\centering
\subfloat[conn\_bench\_sonar\_mines\_rocks]{\includegraphics[scale=0.4]{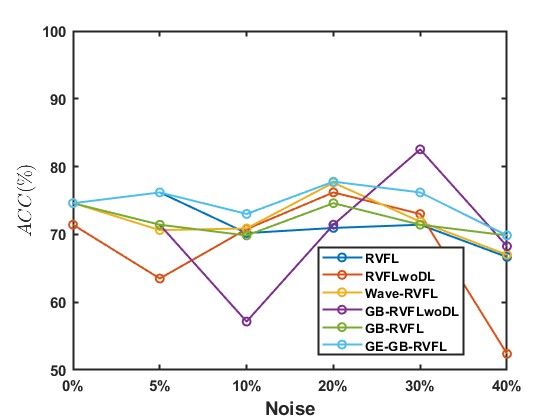}}
\end{minipage}
\begin{minipage}{.5\linewidth}
\centering
\subfloat[ecoli-0-1-4-6\_vs\_5]{\includegraphics[scale=0.4]{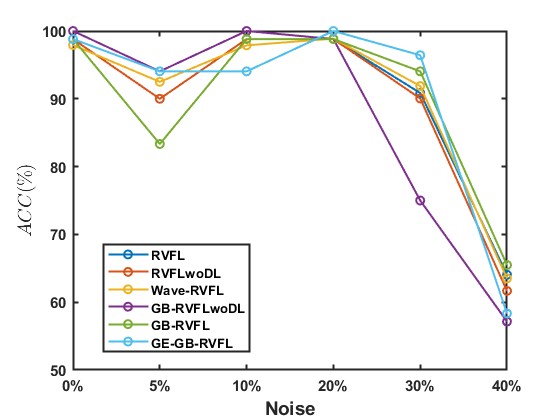}}
\end{minipage}
\par\medskip
\begin{minipage}{.5\linewidth}
\centering
\subfloat[statlog\_heart]{\includegraphics[scale=0.4]{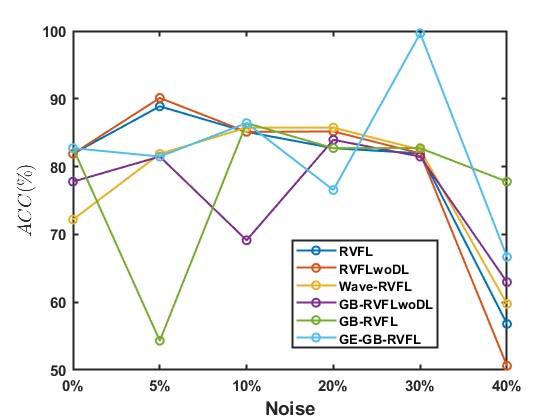}}
\end{minipage}
\begin{minipage}{.5\linewidth}
\centering
\subfloat[yeast-0-2-5-7-9\_vs\_3-6-8]{\includegraphics[scale=0.4]{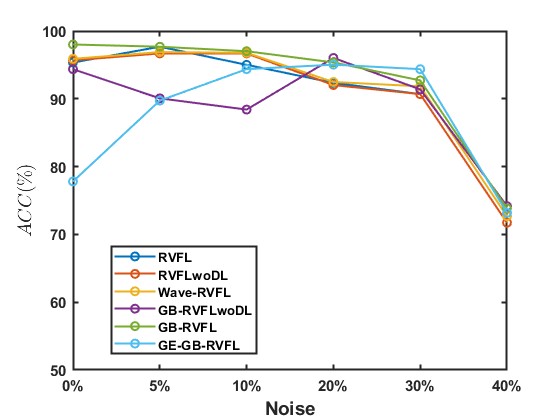}}
\end{minipage}
\caption{Effect of different labels of noise on the performance of the proposed GB-RVFL and GE-GB-RVFL models.}
\label{fig:label_noise_rvfl}
\end{figure*}
\begin{figure*}[ht!]
\begin{minipage}{.315\linewidth}
\centering
\subfloat[bank]{\includegraphics[scale=0.25]{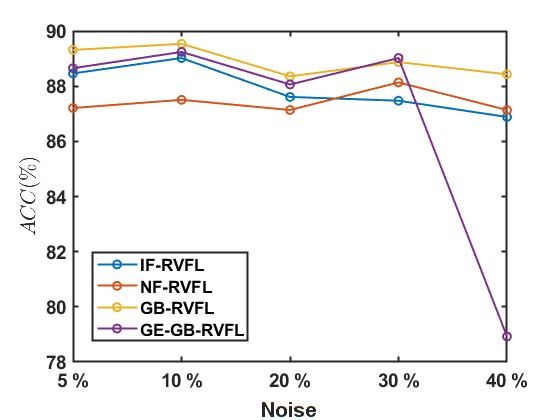}}
\end{minipage}
\begin{minipage}{.315\linewidth}
\centering
\subfloat[fertility]{\includegraphics[scale=0.25]{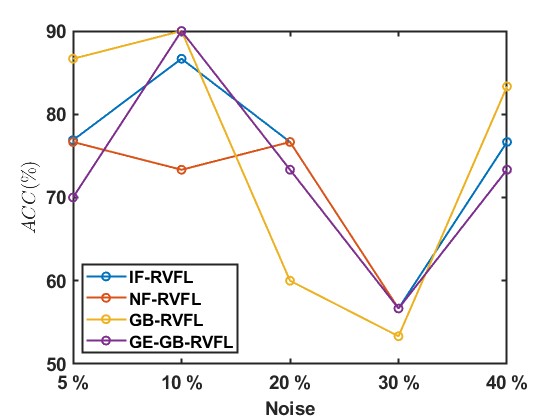}}
\end{minipage}
\begin{minipage}{.315\linewidth}
\centering
\subfloat[ionosphere]{\includegraphics[scale=0.25]{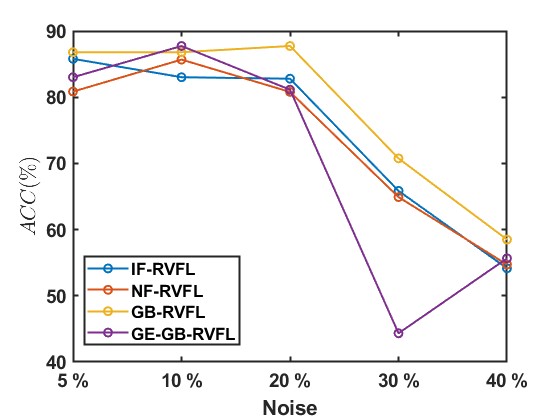}}
\end{minipage}
\caption{Effect of different labels of noise on the performance of the proposed GB-RVFL and GE-GB-RVFL model with the baseline IF-RVFL and NF-RVFL models.}
\label{fig:Noise_labels_NF_RVFL}
\end{figure*}

Table \ref{tab:Noise} and Fig.  \ref{fig:label_noise_rvfl} present experimental results and graphs, respectively, across different noise levels. The proposed GB-RVFL and GE-GB-RVFL models consistently outperform baseline models, showcasing superior accuracy. In each noise level, the proposed GB-RVFL and GE-GB-RVFL emerge as best performers, achieving average accuracies (overall) of $85.7265$ and $85.2408$, respectively. 

Compared with the proposed models, the GB-RVFLwoDL model lags with an overall average accuracy of $81.5653$. This indicates a significant performance gap of approximately $4\%$ between the GB-RVFLwoDL and the proposed models. This emphasizes the advanced noise resilience exhibited by the proposed GB-RVFL and GE-GB-RVFL models. Additionally, the presence of at least one proposed model (sometimes both) among the top two performers in each dataset underscores the robustness of these models due to the fusion of GB computation in handling noise.  
\subsubsection{Comparision among proposed and fuzzy based models} The robustness of fuzzy-based models is widely acknowledged \cite{malik2022alzheimer}, prompting us to evaluate the robustness of our proposed models against state-of-the-art baseline robust models, namely IF-RVFL and NF-RVFL. We conduct our analysis using three datasets: bank, fertility, and ionosphere, with corresponding graphs illustrated in Fig.  \ref{fig:Noise_labels_NF_RVFL}. Notably, our findings consistently demonstrate the robust performance of the proposed GB-RVFL models compared to baseline models. Specifically, when subjected to increasing noise percentage levels in the bank dataset, the proposed GB-RVFL model maintains its accuracy, displaying insensitivity to noise fluctuations. Similarly, for the fertility and ionosphere datasets, the GB-RVFL models consistently exhibit the highest accuracy at a noise label of $40\%$, highlighting their superior robustness over state-of-the-art baseline models. The competitive performance of the GE-GB-RVFL model raises intriguing possibilities. One potential explanation for this phenomenon could be that while preserving the intrinsic geometrical structure of the datasets, the model may, to some extent, sacrifice a portion of its robustness. This delicate balance between structural fidelity and robustness could contribute to the GE-GB-RVFL model's ability to achieve competitive results.

Thus, by utilizing the coarse nature of GBs and focusing on their centers, the proposed GB-RVFL and GE-GB-RVFL models effectively capture the core information while reducing susceptibility to noise located further from central data distribution or clusters.
\subsection{Results and Discussions on
NDC Dataset} \label{subsec:NDC}
We conducted extensive experiments on large datasets generated using David Musicant's NDC Data Generator \cite{ndcdata1998}, ranging from $50$ thousand to $100$ million data samples. For example, NDC-50K and NDC-1M indicate that the dataset comprises $50,000$ and $1$ million data samples, respectively. The results in Table \ref{tab:NDC_accuracy} provide a comprehensive view of the performance of the proposed GB-RVFL and GE-GB-RVFL models in comparison to the baseline RVFL and RVFLwoDL models. The best hyperparameter settings for all the compared models on NDC datasets are reported in Table S.4 of the supplementary material.
\begin{table}
\centering
\caption{Accuracies of the proposed models versus the baseline models on NDC datasets.}
\label{tab:NDC_accuracy}
\resizebox{\textwidth}{!}{
\begin{tabular}{l|ccccc}\hline \vspace{-2mm}\\ 
 \text{Model} $\rightarrow$
                           & RVFL \cite{pao1994learning}                 & RVFLwoDL  \cite{huang2006extreme}                  & GB-RVFLwoDL$^{\bigstar}$                 & GB-RVFL$^{\bigstar}$                 & GE-GB-RVFL$^{\bigstar}$   
                           \vspace{1mm}\\ \hline \vspace{-2mm}\\ 
Dataset $\downarrow$           & ACC        & ACC         & ACC         & ACC         & ACC 
                           \vspace{1mm}\\ \hline \vspace{-2mm}\\
NDC-50K  & 80.9435              & 80.1134              & 82.1687 & 84.4884          & \textbf{85.3785} \\
NDC-100K & 78.2968              & 78.6267              & 88.2494 & \textbf{88.4044} & 83.8742          \\
NDC-500K & 82.3814              & 81.1474              & 80.7758 & 83.2042          & \textbf{85.3393} \\
NDC-1M   & 81.6562              & 81.5717              & 83.1739 & 85.2399          & \textbf{86.1404} \\
NDC-3M   & 80.8727              & 80.4434              & 80.7375 & 82.6701          & \textbf{83.9588} \\
NDC-5M   & 84.0435              & 83.608               & 83.9787 & 84.8743          & \textbf{85.3805} \\
NDC-10M  & 82.8764              & 82.7896              & 86.1019 & \textbf{86.6179} & 85.7042          \\
NDC-30M  & a & a & 82.6621 & 83.4257          & \textbf{83.4441} \\
NDC-50M  & a & a & 80.9622 & 83.5944          & \textbf{83.7941} \\
NDC-100M & a & a & 80.9646 & 84.4071          & \textbf{85.0702}  \vspace{1mm}\\ \hline \vspace{-3mm}\\
Average ACC & 81.5815              & 81.1857              & 82.9775 & 84.6926          & \textbf{84.8084}\\ \hline
\multicolumn{6}{l}{The best-performing model in terms of accuracy is indicated by boldface in each row.}\\
\multicolumn{6}{l}{``a" denotes the code terminated due to out of memory.  ACC is an acronym for accuracy.}\\
\multicolumn{6}{l}{$^{\bigstar}$ denotes the proposed models.}\\
\end{tabular}%
}
\end{table}
Analysis of Table \ref{tab:NDC_accuracy} reveals that the GE-GB-RVFL model outperforms as the top performer, achieving the highest average accuracy at $84.8084\%$, followed closely by GB-RVFL with $84.6926\%$ average accuracy. Notably, RVFL and RVFLwoDL encounter memory issues with larger datasets (e.g., NDC-30M, NDC-50M and NDC-100M), while our proposed GB-based models handle these datasets seamlessly. This emphasizes the adept adaptation and leveraging of the GB and GE frameworks by our proposed GB-RVFL and GE-GB-RVFL models. Furthermore, it attests to the scalability and efficacy exhibited by proposed GB-RVFL and GE-GB-RVFL models, demonstrating a notable capacity to generalize effectively, especially when confronted with large and intricate datasets such as NDC. Scalability in the proposed GB-RVFL and GE-GBRVFL models stems from their training process, which employs GBs instead of the entire dataset. Because the number of GBs is much smaller than the total data points, the models' scalability is improved.
\subsection{Application on Biomedical Domain}
To show the applicability of the proposed GB-RVFL and GE-GB-RVFL models in real-world scenarios, we employ them in biomedical datasets, namely, the BreakHis dataset (for breast cancer classification) and ADNI datasets (for AD classification). To thoroughly evaluate the effectiveness of the proposed GB-RVFL and GE-GB-RVFL models, we assess their performance across multiple metrics, including Accuracy (ACC), Sensitivity, Specificity, and Precision. Detailed formulas for all the metrics can be found in Section S.I of the supplementary material.
\begin{table*}[ht!]
\centering
\caption{Accuracies of the proposed GB-RVFL and GE-GB-RVFL models against the baseline models on Biomedical datasets.}
\label{tab:Biomedical datasets}
\resizebox{1.0\textwidth}{!}{
\begin{tabular}{lcccccc}
\hline
Dataset & RVFL \cite{pao1994learning} & RVFLwoDL \cite{huang2006extreme} & IF-RVFL \cite{malik2022alzheimer}  & GB-RVFLwoDL$^{\bigstar}$ & GB-RVFL$^{\bigstar}$ & GE-GB-RVFL$^{\bigstar}$ \\
 & (ACC, Sensitivity) & (ACC, Sensitivity) & (ACC, Sensitivity) & (ACC, Sensitivity) & (ACC, Sensitivity) & (ACC, Sensitivity) \\
 & (Specificity, Precision) & (Specificity, Precision) & (Specificity, Precision) & (Specificity, Precision) & (Specificity, Precision) & (Specificity, Precision) \\ \hline
adenosis\_vs\_ductal\_carcinoma & $(85.26, 76.67)$ & $(77.89, 63.33)$ & $(88.42, 80)$ & $(85.26, 86.67)$ & $(91.58, 92.59)$ & $(81.05, 60)$ \\
 & $(89.23, 76.67)$ & $(84.62, 65.52)$ & $(92.31, 82.76)$ & $(84.62, 72.22)$ & $(91.18, 80.65)$ & $(90.77, 75)$ \\
adenosis\_vs\_lobular\_carcinoma & $(54.79, 42.86)$ & $(46.58, 42.86)$ & $(52.05, 0)$ & $(50.68, 42.86)$ & $(53.42, 17.14)$ & $(58.9, 45.71)$ \\
 & $(65.79, 53.57)$ & $(50, 44.12)$ & $(100, 0)$ & $(57.89, 48.39)$ & $(86.84, 54.55)$ & $(71.05, 59.26)$ \\
adenosis\_vs\_mucinous\_carcinoma & $(66.27, 18.18)$ & $(60.24, 0)$ & $(61.45, 15.15)$ & $(67.47, 45.45)$ & $(57.83, 54.55)$ & $(61.45, 43.75)$ \\
 & $(98, 85.71)$ & $(100, 0)$ & $(92, 55.56)$ & $(82, 62.5)$ & $(60, 47.37)$ & $(72.55, 50)$ \\
adenosis\_vs\_papillary\_carcinoma & $(55.41, 42.42)$ & $(52.7, 36.36)$ & $(55.41, 0)$ & $(54.05, 60.61)$ & $(51.35, 37.93)$ & $(55.41, 40)$ \\
 & $(65.85, 50)$ & $(65.85, 46.15)$ & $(100, 0)$ & $(48.78, 48.78)$ & $(60, 37.93)$ & $(69.23, 53.85)$ \\
CN\_vs\_AD & $(88, 85.96)$ & $(84, 80.7)$ & $(77.6, 73.68)$ & $(79.2, 64.91)$ & $(84, 73.68)$ & $(85.6, 80.7)$ \\
 & $(89.71, 87.5)$ & $(86.76, 83.64)$ & $(80.88, 76.36)$ & $(91.18, 86.05)$ & $(92.65, 89.36)$ & $(89.71, 86.79)$ \\
CN\_vs\_MCI & $(72.87, 86.78)$ & $(73.4, 86.78)$ & $(73.94, 86.78)$ & $(73.4, 86.78)$ & $(70.21, 87.6)$ & $(75, 87.6)$ \\
 & $(47.76, 75)$ & $(49.25, 75.54)$ & $(50.75, 76.09)$ & $(49.25, 75.54)$ & $(38.81, 72.11)$ & $(52.24, 76.81)$ \\
fibroadenoma\_vs\_ductal\_carcinoma & $(79.85, 77.92)$ & $(79.1, 77.92)$ & $(79.85, 75.32)$ & $(70.9, 66.23)$ & $(79.85, 79.22)$ & $(78.36, 76.62)$ \\
 & $(82.46, 85.71)$ & $(80.7, 84.51)$ & $(85.96, 87.88)$ & $(77.19, 79.69)$ & $(80.7, 84.72)$ & $(80.7, 84.29)$ \\
fibroadenoma\_vs\_lobular\_carcinoma & $(65.49, 100)$ & $(64.6, 98.65)$ & $(65.49, 100)$ & $(49.56, 64.86)$ & $(66.37, 83.78)$ & $(60.18, 68.92)$ \\
 & $(0, 65.49)$ & $(0, 65.18)$ & $(0, 65.49)$ & $(20.51, 60.76)$ & $(33.33, 70.45)$ & $(43.59, 69.86)$ \\
fibroadenoma\_vs\_mucinous\_carcinoma & $(50, 52.44)$ & $(51.64, 58.54)$ & $(56.56, 69.51)$ & $(59.84, 58.54)$ & $(55.74, 57.89)$ & $(61.48, 64.63)$ \\
 & $(45, 66.15)$ & $(37.5, 65.75)$ & $(30, 67.06)$ & $(62.5, 76.19)$ & $(52.17, 66.67)$ & $(55, 74.65)$ \\
fibroadenoma\_vs\_papillary\_carcinoma & $(61.95, 77.63)$ & $(55.75, 76.32)$ & $(59.29, 72.37)$ & $(53.98, 73.68)$ & $(64.6, 93.42)$ & $(66.37, 97.37)$ \\
 & $(29.73, 69.41)$ & $(13.51, 64.44)$ & $(32.43, 68.75)$ & $(13.51, 63.64)$ & $(5.41, 66.98)$ & $(2.7, 67.27)$ \\
MCI\_vs\_AD & $(68.18, 29.51)$ & $(69.89, 32.79)$ & $(67.05, 24.59)$ & $(67.05, 19.67)$ & $(69.32, 49.18)$ & $(72.16, 39.34)$ \\
 & $(88.7, 58.06)$ & $(89.57, 62.5)$ & $(89.57, 55.56)$ & $(92.17, 57.14)$ & $(80, 56.6)$ & $(89.57, 66.67)$ \\
phyllodes\_tumour\_vs\_ductal\_carcinoma & $(84.54, 85)$ & $(81.44, 92.5)$ & $(84.54, 82.5)$ & $(60.82, 20)$ & $(83.51, 88.57)$ & $(90.72, 94.12)$ \\
 & $(84.21, 79.07)$ & $(73.68, 71.15)$ & $(85.96, 80.49)$ & $(89.47, 57.14)$ & $(80.65, 72.09)$ & $(88.89, 82.05)$ \\
phyllodes\_tumour\_vs\_lobular\_carcinoma & $(56.58, 57.14)$ & $(48.68, 48.57)$ & $(57.89, 34.29)$ & $(50, 37.14)$ & $(59.21, 42.86)$ & $(59.21, 71.43)$ \\
 & $(56.1, 52.63)$ & $(48.78, 44.74)$ & $(78.05, 57.14)$ & $(60.98, 44.83)$ & $(73.17, 57.69)$ & $(48.78, 54.35)$ \\
phyllodes\_tumour\_vs\_mucinous\_carcinoma & $(47.67, 55.56)$ & $(51.16, 66.67)$ & $(50, 44.44)$ & $(48.84, 62.96)$ & $(53.49, 24.32)$ & $(36.05, 48.15)$ \\
 & $(44.07, 31.25)$ & $(44.07, 35.29)$ & $(52.54, 30)$ & $(42.37, 33.33)$ & $(75.51, 42.86)$ & $(30.51, 24.07)$ \\
phyllodes\_tumour\_vs\_papillary\_carcinoma & $(44.74, 34.29)$ & $(60.53, 45.71)$ & $(44.74, 11.43)$ & $(60.53, 22.86)$ & $(56.58, 57.14)$ & $(53.95, 41.67)$ \\
 & $(53.66, 38.71)$ & $(73.17, 59.26)$ & $(73.17, 26.67)$ & $(92.68, 72.73)$ & $(56.1, 52.63)$ & $(65, 51.72)$ \\
tubular\_adenoma\_vs\_ductal\_carcinoma & $(68.63, 53.19)$ & $(67.65, 51.06)$ & $(67.65, 46.81)$ & $(66.67, 40.43)$ & $(66.67, 42.55)$ & $(67.65, 40.43)$ \\
 & $(81.82, 71.43)$ & $(81.82, 70.59)$ & $(85.45, 73.33)$ & $(89.09, 76)$ & $(87.27, 74.07)$ & $(90.91, 79.17)$ \\
tubular\_adenoma\_vs\_lobular\_carcinoma & $(45.68, 33.33)$ & $(56.79, 47.62)$ & $(51.85, 45.24)$ & $(45.68, 57.14)$ & $(56.79, 40.54)$ & $(50.62, 23.08)$ \\
 & $(58.97, 46.67)$ & $(66.67, 60.61)$ & $(58.97, 54.29)$ & $(33.33, 48)$ & $(70.45, 53.57)$ & $(76.19, 47.37)$ \\
tubular\_adenoma\_vs\_mucinous\_carcinoma & $(50, 23.26)$ & $(52.22, 23.26)$ & $(53.33, 18.6)$ & $(57.78, 23.26)$ & $(63.33, 17.95)$ & $(61.11, 43.59)$ \\
 & $(74.47, 45.45)$ & $(78.72, 50)$ & $(85.11, 53.33)$ & $(89.36, 66.67)$ & $(98.04, 87.5)$ & $(74.51, 56.67)$ \\
tubular\_adenoma\_vs\_papillary\_carcinoma & $(55.56, 41.46)$ & $(49.38, 70.73)$ & $(54.32, 21.95)$ & $(40.74, 39.02)$ & $(60.49, 46.34)$ & $(46.91, 89.47)$ \\
 & $(70, 58.62)$ & $(27.5, 50)$ & $(87.5, 64.29)$ & $(42.5, 41.03)$ & $(75, 65.52)$ & $(9.3, 46.58)$ \\ \hline
Average ACC & $63.23$ & $62.3$ & $63.23$ & $60.13$ & $\mathbf{65.49}$ & $64.32$ \\ \hline
Average Rank & $3.37$ & $4$ & $3.45$ & $4.55$ & $2.95$ & $\mathbf{2.68}$ \\ \hline
\multicolumn{7}{l}{The best-performing model in terms of average accuracy and average ranks are indicated by boldface in the second last and last rows, respectively. ACC is an acronym for accuracy.}\\ \multicolumn{7}{l}{$^{\bigstar}$ denotes the proposed models.}
\end{tabular}}
\end{table*}
\begin{figure*}
\begin{minipage}{.315\linewidth}
\centering
\subfloat[CN\_vs\_AD]{\includegraphics[scale=0.25]{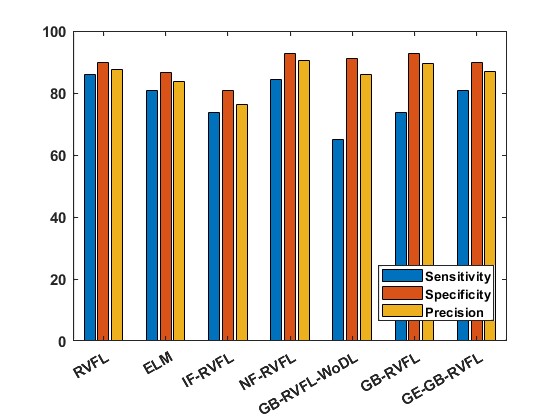}}
\end{minipage}
\begin{minipage}{.315\linewidth}
\centering
\subfloat[CN\_vs\_MCI]{\includegraphics[scale=0.25]{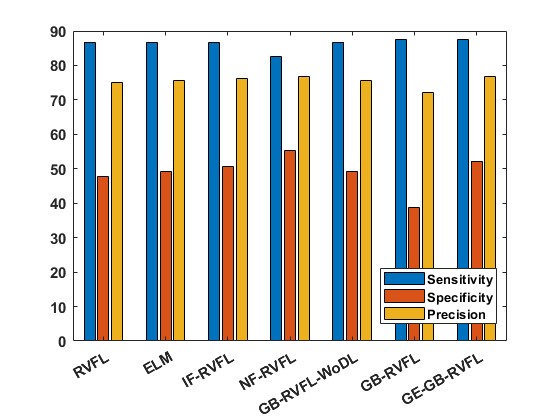}}
\end{minipage}
\begin{minipage}{.315\linewidth}
\centering
\subfloat[MCI\_vs\_AD]{\includegraphics[scale=0.25]{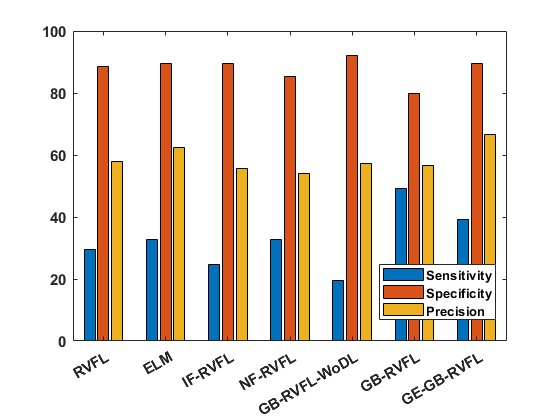}}
\end{minipage}
\caption{Performance comparison of the proposed GB-RVFL and GE-GB-RVFL models against baseline models on ADNI data using sensitivity, specificity, and precision.}
\label{Bar graph of specificity, sensitivity, and precision on ADNI dataset.}
\end{figure*}
\begin{itemize}
    \item \textbf{\textit{BreakHis dataset}}: BreakHis dataset comprises histopathology scans, as detailed in reference \cite{spanhol2015dataset}. Our study included 1240 scans, all at 400 times magnification. These scans primarily fell into two categories: malignant and benign. Among the malignant scans, there were subclasses such as lobular carcinoma, ductal carcinoma, papillary carcinoma and mucinous carcinoma with respective scan counts of 137, 208, 138, and 169. Similarly, the benign class had subcategories like tubular adenoma, phyllodes tumour, adenosis, and fibroadenoma with corresponding scan counts of 130, 115, 106, and 237. Our approach for feature extraction from these histopathological scans mirrored the methodology outlined in the references \cite{KUMARI2024111454}. We employed a total of 16 combinations involving both benign and malignant classes to differentiate between the various subclasses of benign and malignant cancers, as summarized in Table \ref{tab:Biomedical datasets}.
    \item \textbf{\textit{ADNI dataset}}: Alzheimer's disease (AD) stands out as a prevalent neurodegenerative condition \cite{tanveer2024ensemble}, accounting for approximately 70\% of dementia cases, according to \citet{khojaste2022deep}. Despite ongoing research efforts, the exact cause of AD remains elusive, and an effective cure is yet to be discovered \cite{tanveer2023deep}. This underscores the critical need for early detection methods. In our study, we utilized scans from the ADNI dataset, accessible via \textit{\href{}{https://adni.loni.usc.edu/}}. The ADNI project was initiated with the aim to investigate neuroimaging modalities like positron emission tomography (PET), magnetic resonance imaging (MRI), and other diagnostic assessments for AD, particularly focusing on mild cognitive impairment (MCI) stages. Our approach for feature extraction aligns with the methodology described in reference \cite{ganaie2024graph}. The dataset we worked with encompasses three primary comparisons: MCI versus AD (MCI\_vs\_AD), control normal (CN) versus MCI (CN\_vs\_MCI), and CN versus AD (CN\_vs\_AD).
\end{itemize}

\subsubsection{\textbf{Results and Discussion}} Table \ref{tab:Biomedical datasets} presents the accuracy, sensitivity, specificity, and precision values for both the proposed models and the baseline models. Further details on the best hyperparameters can be found in Table S.5 of the supplementary material. Our analysis demonstrates that among all the models evaluated, the proposed GB-RVFL achieves the highest accuracy at $65.49\%$, followed by the GE-GB-RVFL with an accuracy of $64.32\%$, ranking as the second-highest. From the last row of the table, we observe that the average rank of the proposed GE-GB-RVFL is minimal, followed by GB-RVFL. This also indicates the superior performance of our proposed models.
For a more nuanced understanding of performance, we examined sensitivity, specificity, and precision through graphs plotted with respect to the ADNI datasets, as depicted in Fig.  \ref{Bar graph of specificity, sensitivity, and precision on ADNI dataset.}. As noted in the literature \cite{tanveer2020machine}, distinguishing the MCI\_vs\_AD case poses a significant challenge. However, our findings indicate that the proposed GB-RVFL and GE-GB-RVFL models showcase the highest and second-highest sensitivity levels, a crucial metric in biomedical data analysis. This trend is similarly observed in the CN\_vs\_MCI case. In summary, our proposed models demonstrate robust performance across various metrics, showcasing their efficacy in real-world datasets, i.e., breast cancer and AD classification.
\section{Interpretability of the Proposed Models Through Graph Embedding and Granular Computing}
\label{interpretability}
A central claim of this paper is that the proposed GE-GB-RVFL model maintains the geometric structure of the original data, an aspect that randomized models often struggle to preserve. This ability to leverage the inherent geometry of the data is particularly valuable, as it can lead to more informed and accurate decision-making. Thus, ensuring that the model makes decisions based on the actual features of the dataset—rather than functioning as a black-box model—becomes even more critical. In this section, we provide both a theoretical framework and visual evidence to demonstrate that the interpretability of the proposed GE-GB-RVFL and GB-RVFL models is superior to that of traditional baseline models. By doing so, we aim to show not only how the model performs but also why it arrives at its decisions. We begin by exploring which features each model relies on when making its final predictions, offering insights into their decision-making processes.

For GE-GB-RVFL, from Eq. \eqref{eq:GE-GB-RVFL_weights}, we observe that the weight matrix is defined as: $\Omega= \left(D^{t}D + \frac{1}{\mathcal{C}}I + \frac{\alpha}{\mathcal{C}}U\right)^{-1}D^{t}\mathcal{W}.$
This implies that the features used to make the final decision are influenced by the inverse of the matrix $\left(D^{t}D + \frac{1}{\mathcal{C}}I + \frac{\alpha}{\mathcal{C}}U\right)^{-1}D^{t}$, i.e., $(D^{t})^\dagger\left(D^{t}D + \frac{1}{\mathcal{C}}I + \frac{\alpha}{\mathcal{C}}U\right) \eqqcolon \mathcal{E}_1 \in \mathbb{R}^{(P+g)\times k}$, where $\dagger$ is the Moore-Penrose inverse. Furthermore, for the GB-RVFL model (from \eqref{eq:Omega_GB_RVFL}), the features used to predict the final output can be expressed as $(D^{t})^\dagger\left(D^{t}D + \frac{1}{\mathcal{C}}I\right) \eqqcolon \mathcal{E}_2 \in \mathbb{R}^{(P+g)\times k}$. Similarly, for the features related to the original GB matrix $\mathcal{O}$, the corresponding feature matrix can be defined as $(\mathcal{O}^{t})^\dagger\left(\mathcal{O}^{t}\mathcal{O} + \frac{1}{\mathcal{C}}I\right) \eqqcolon \mathcal{E}_3 \in \mathbb{R}^{P\times M}$.

For the RVFL, using Eq. \eqref{eq:pseudo-inverse}, the output weights $\mathcal{Q}$ are defined as: $\mathcal{Q}=[V \oplus \mathcal{G}]^{\dagger}Z= \left([V \oplus \mathcal{G}]^{t}[V \oplus \mathcal{G}] + \frac{1}{\mathcal{C}}I\right)^{-1}[V \oplus \mathcal{G}]^{t} Z$. Thus, the features contributing to the final decision of the RVFL can be represented as: $([V \oplus \mathcal{G}]^{t})^\dagger\left([V \oplus \mathcal{G}]^{t}[V \oplus \mathcal{G}] + \frac{1}{\mathcal{C}}I\right) \eqqcolon \mathcal{E}_4 \in \mathbb{R}^{(P+g)\times M}$. Similarly, the features associated with the RVFLwoDL and for the original input features are given by: $(\mathcal{G}^{t})^\dagger\left(\mathcal{G}^{t}\mathcal{G} + \frac{1}{\mathcal{C}}I\right) \eqqcolon \mathcal{E}_5 \in \mathbb{R}^{g\times M}$ and $(V^{t})^\dagger\left(V^{t}V + \frac{1}{\mathcal{C}}I\right) \eqqcolon \mathcal{E}_6 \in \mathbb{R}^{P\times M}$, respectively.

We adjust the dimensions of each \(\mathcal{E}_i\) (for \(i=1,2\)) to match that of \(\mathcal{E}_3\) by cropping the extra rows and/or columns. Similarly, we modified the dimensions of each \(\mathcal{E}_i\) (for \(i=4,5\)) to align with \(\mathcal{E}_6\) through the same cropping process. These modified matrices are denoted as \(\mathcal{E}_i'\) (for \(i=1,2,4,5\)). For consistency, we also renamed \(\mathcal{E}_3\) and \(\mathcal{E}_6\) as \(\mathcal{E}_3'\) and \(\mathcal{E}_6'\), respectively.

\begin{table}[]
\centering
\caption{Feature interpretability results for the datasets \text{conn\_bench\_sonar\_mines\_rocks} and \text{acute\_inflammation}, based on the Frobenius norm distance of the feature distance matrices.}
\label{tab:interpretability}
\resizebox{12cm}{!}{%
\begin{tabular}{lcccc}
\\ \hline
Dataset $\downarrow$ / \text{Distance} $\rightarrow$    & $\mathcal{DDE}_1$    & $\mathcal{DDE}_2$    & $\mathcal{DDE}_4$    & $\mathcal{DDE}_5$    \\ \hline
acute\_inflammation              & $3.49E-05$ & $3.60E-05$ & $0.2559$   & $92.3832$   \\
conn\_bench\_sonar\_mines\_rocks & $6.03E-05$ & $6.13E-05$ & $7.79E-05$ & $2186.1229$ \\ \hline
\end{tabular}%
}
\end{table}

The next step involves calculating the pairwise distance between each row of the feature matrices \(\mathcal{E}_i'\) (for \(i = 1, 2, \ldots, 6\)), denoted as \(\mathcal{DE}_i\), where each entry represents the distance between features within the matrix. Note that $\mathcal{DE}_i \in \mathbb{R}^{k\times k}$ for $i=1,2,3$ and $\mathcal{DE}_i \in \mathbb{R}^{M\times M}$ for $i=4,5,6$. Once we obtain feature distance matrices \(\mathcal{DE}_i\), we can compute the distance between each feature distance matrix \(\mathcal{DE}_i\) (for \(i = 1, 2\)) and the feature distance matrix \(\mathcal{DE}_3\) using the Frobenius norm. This is defined as:

\[
\mathcal{DDE}_i = \|\mathcal{DE}_i - \mathcal{DE}_3\|_F, \quad \text{for} \quad i = 1, 2,
\]

Similarly, we define \(\mathcal{DDE}_4\) and \(\mathcal{DDE}_5\) as:

\[
\mathcal{DDE}_i = \|\mathcal{DE}_i - \mathcal{DE}_6\|_F, \quad \text{for} \quad i = 4, 5,
\]

where \(\|\cdot\|_F\) denotes the Frobenius norm, which provides a measure of the difference between the distance matrices.

The core concept is that smaller values of \(\mathcal{DDE}_i\) signify a closer alignment between the features in \(\mathcal{E}_i\) and the original features (\(V\)) or the original GB features (\(\mathcal{O}\)). Therefore, lower \(\mathcal{DDE}_i\) values reflect superior preservation of the original feature structure within the learned representations of the model. This suggests that the model effectively captures and retains the essential characteristics of the original data, enhancing interpretability and reliability in decision-making processes.



\begin{figure*}[ht!]
\begin{minipage}{.33\linewidth}
\centering
\subfloat[$\mathcal{DE}_1$]{\includegraphics[scale=0.3]{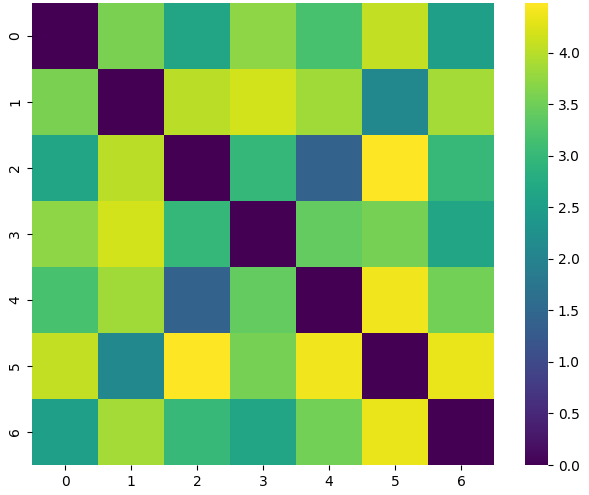}}
\end{minipage}
\begin{minipage}{.33\linewidth}
\centering
\subfloat[$\mathcal{DE}_2$]{\includegraphics[scale=0.3]{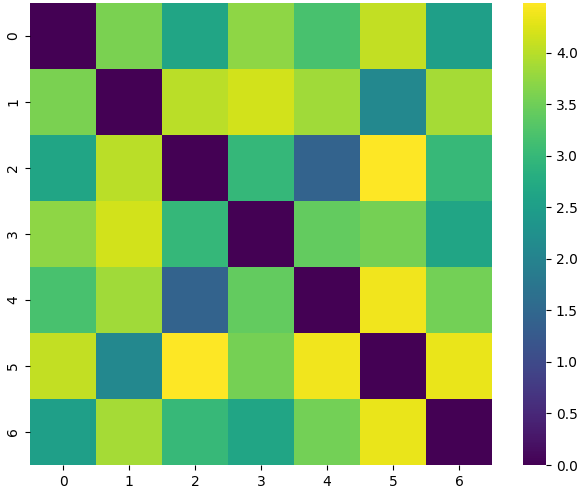}}
\end{minipage}
\begin{minipage}{.33\linewidth}
\centering
\subfloat[$\mathcal{DE}_3$]{\includegraphics[scale=0.3]{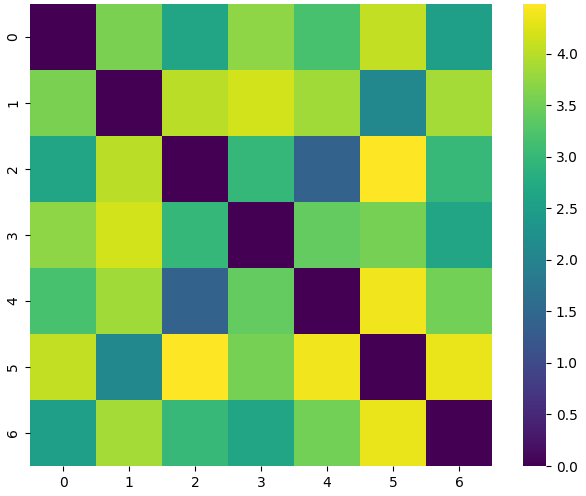}}
\end{minipage}
\par\medskip
\begin{minipage}{.33\linewidth}
\centering
\subfloat[$\mathcal{DE}_4$]{\includegraphics[scale=0.3]{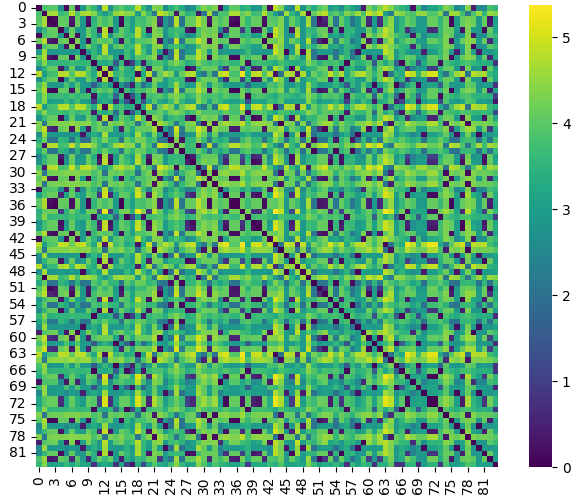}}
\end{minipage}
\begin{minipage}{.33\linewidth}
\centering
\subfloat[$\mathcal{DE}_5$]{\includegraphics[scale=0.3]{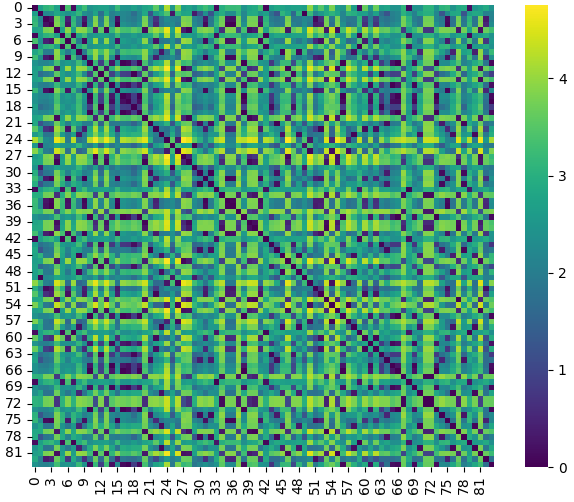}}
\end{minipage}
\begin{minipage}{.33\linewidth}
\centering
\subfloat[$\mathcal{DE}_6$]{\includegraphics[scale=0.3]{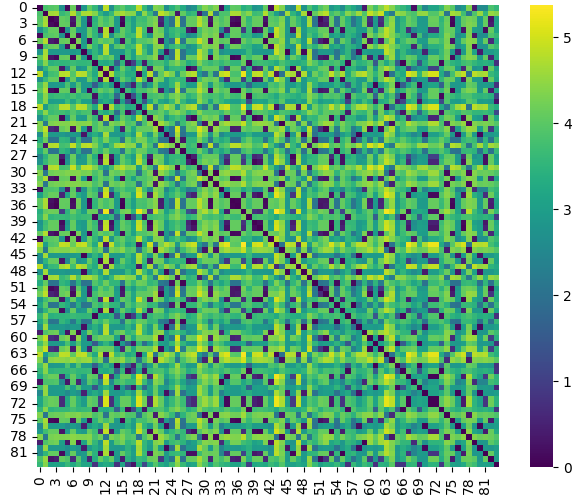}}
\end{minipage}
\caption{Visualization of the feature distance matrices $\mathcal{DE}_i$ through Heat map on ``\text{acute\_inflammation}'' dataset.}
\label{fig:heatmap_acute}
\end{figure*}
\begin{figure*}[ht!]
\begin{minipage}{.33\linewidth}
\centering
\subfloat[$\mathcal{DE}_1$]{\includegraphics[scale=0.3]{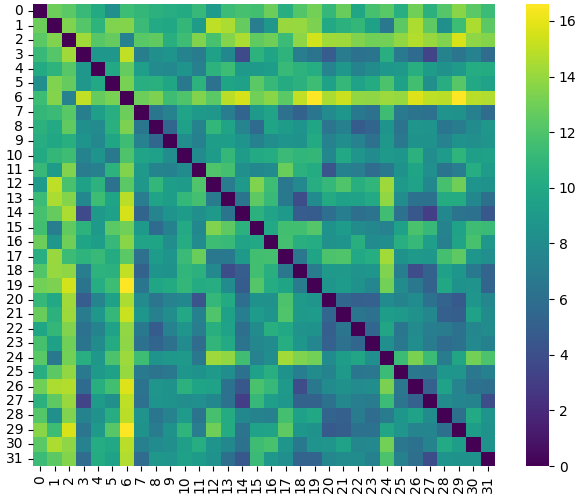}}
\end{minipage}
\begin{minipage}{.33\linewidth}
\centering
\subfloat[$\mathcal{DE}_2$]{\includegraphics[scale=0.3]{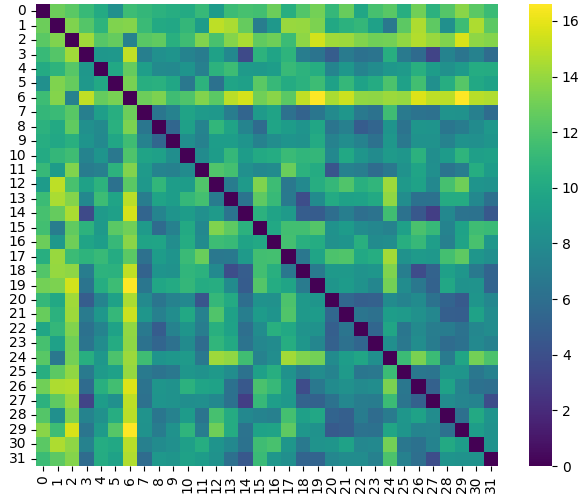}}
\end{minipage}
\begin{minipage}{.33\linewidth}
\centering
\subfloat[$\mathcal{DE}_3$]{\includegraphics[scale=0.3]{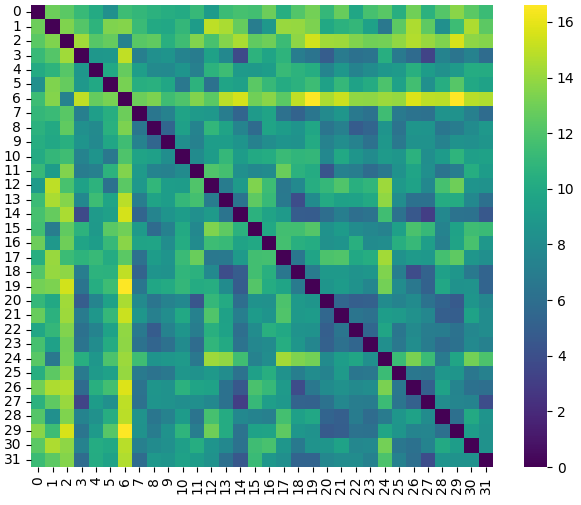}}
\end{minipage}
\par\medskip
\begin{minipage}{.33\linewidth}
\centering
\subfloat[$\mathcal{DE}_4$]{\includegraphics[scale=0.3]{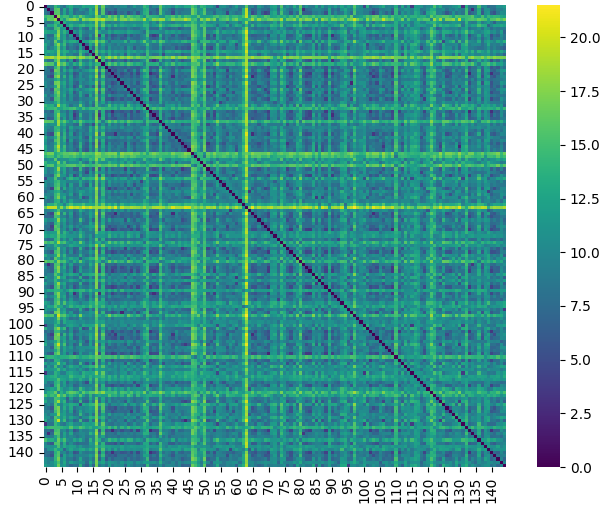}}
\end{minipage}
\begin{minipage}{.33\linewidth}
\centering
\subfloat[$\mathcal{DE}_5$]{\includegraphics[scale=0.3]{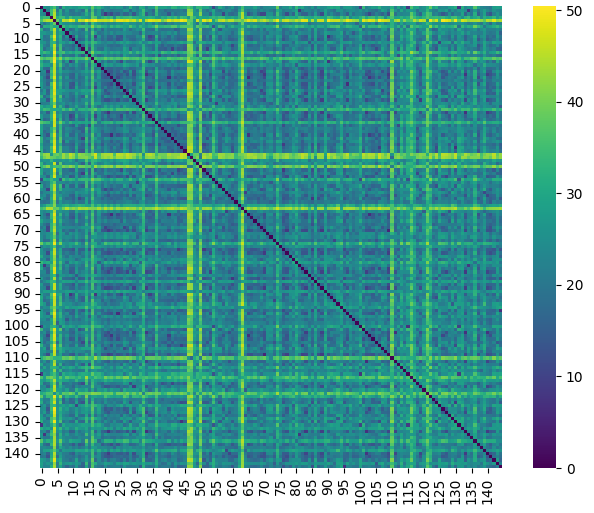}}
\end{minipage}
\begin{minipage}{.33\linewidth}
\centering
\subfloat[$\mathcal{DE}_6$]{\includegraphics[scale=0.3]{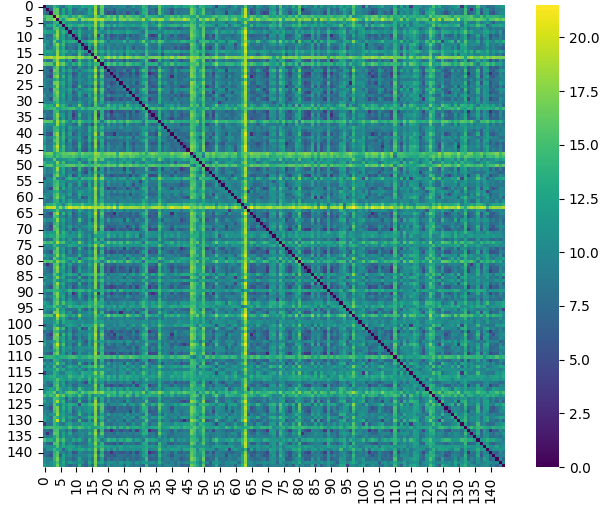}}
\end{minipage}
\caption{Visualization of the feature distance matrices $\mathcal{DE}_i$ through Heat map on ``\text{conn\_bench\_sonar\_mines\_rocks}'' dataset.}
\label{fig:heatmap_conn}
\end{figure*}

To assess the feature interpretability of our proposed models, we conducted experiments on two datasets: ``\text{conn\_bench\_sonar\_mines\_rocks}'' and ``\text{acute\_inflammation}''. The results are summarized in Table \ref{tab:interpretability}. Notably, the distance \(\mathcal{DDE}_1\) is the smallest among the measured values, indicating that the features associated with the GE-GB-RVFL model are more closely aligned with the original GB features. This alignment is attributed to the use of graph embedding, which effectively preserves the geometric relationships inherent in the data, along with a slight contribution from granular computing techniques.

Following this, the GB-RVFL model shows a slightly larger distance, reinforcing the enhanced feature interpretability of our proposed model. Furthermore, the distance associated with RVFL features is smaller than that of the RVFLwoDL features, highlighting the significance of direct links in feature representation.

These findings demonstrate that the feature alignment of the proposed models is significantly improved—by factors ranging from \(10^4\) to \(10^9\)—compared to the baseline RVFL and RVFLwoDL models. This enhancement implies a greater degree of feature interpretability for both proposed models.

Additionally, we visualize the distances among the features of different matrices \(\mathcal{E}_i\) by generating heat maps for each \(\mathcal{DE}_i\), as depicted in Figs. \ref{fig:heatmap_acute} and \ref{fig:heatmap_conn}. Since the dimension of \(\mathcal{DE}_i\) is influenced by the sample space's dimension, i.e., \(M\) (the number of original samples) and \(k\) (the number of GBs), with \(k\) $<<$ \(M\). Consequently, the dimensions of the heat maps vary accordingly in the figures. These visual representations further illustrate the relationships between features, providing a clearer understanding of how effectively our models capture the original feature structure. By combining theoretical insights with visual evidence, we demonstrate that our proposed GE-GB-RVFL and GB-RVFL models not only outperform traditional baselines but also provide clarity on their decision-making processes, enhancing their interpretability and reliability.
\section{Discussion}
\label{discussion}
This section delves into a fundamental inquiry: do the proposed models achieve a harmonious balance between scalability and accuracy within the realm of granular computing? Subsequently, we conduct various sensitivity analyses of the hyperparameters of the proposed models to assess the effect of various hyperparameters on the performance of the proposed GB-RVFL and GE-GB-RVFL models. These hyperparameters include $pur$ (purity threshold), $num$ (\# minimum GBs), $Actfun$ (activation function), $\mathcal{C}$ (regularization parameter) and graph regularization parameter ($\alpha$).
\subsection{Balance Between Number of GBs Generated and the Resulting Accuracy of the Proposed GB-RVFL and GE-GB-RVFL Models with Different Purities}
Purity significantly impacts the formation of GBs, thus influencing the overall performance of GB-RVFL and GE-GB-RVFL models. Our objective is to explore the correlation between the number of GBs generated ($number(GB)$) and the threshold purity ($pur$), and how this correlation affects our models' performance. We've chosen five diverse UCI and KEEL datasets to showcase the adaptability and effectiveness of GB-RVFL and GE-GB-RVFL.
\begin{table*}[htp!]
\centering
    \caption{The number of GBs and the corresponding Accuracies by the proposed GB-RVFL model under different purities.}
    \label{tab:pur1}
  \resizebox{0.85\textwidth}{!}{
\begin{tabular}{ccccccccc}
\hline
$pur$ & $1$ & $0.97$ & $0.94$ & $0.91$ & $0.88$ & $0.85$ & $0.82$ &  $0.79$ \\ \hline
{Dataset} & ACC & ACC & ACC & ACC & ACC & ACC & ACC   & ACC   \\
 & $number(GB)$ & $number(GB)$ & $number(GB)$ & $number(GB)$ & $number(GB)$ & $number(GB)$ & $number(GB)$  & $number(GB)$ \\ \hline
blood & 78.22 & 76 & 77.33 & 76.44 & 75.11 & 76.89  & 76.89  & 56.89 \\
 & 349 & 354 & 355 & 340 & 314 & 245 & 211 & 172 \\
breast\_cancer & 74.42 & 74.42 & 77.91 & 69.77 & 74.42 & 67.44 & 80.23 & 66.28 \\
 & 139 & 130 & 142 & 132 & 137 & 119 & 119 & 90 \\
breast\_cancer\_wisc\_prog & 63.33 & 68.33 & 60 & 70 & 73.33 & 68.33 & 55&  66.67 \\
 & 94 & 88 & 84 & 88 & 85 & 73 & 71 &  52 \\
chess\_krvkp & 94.99 & 93.95 & 93.85 & 93.64 & 94.16 & 90.51 & 91.14  & 89.68 \\
 & 1168 & 1195 & 1131 & 1062 & 969 & 858 & 757   &  619 \\
cleve & 83.33 & 72.22 & 84.44 & 82.22 & 71.11 & 68.89 & 65.56  &  73.33 \\
 & 108 & 113 & 114 & 100 & 87 & 65 & 45  &  34 \\ \hline
\end{tabular}}
\end{table*}
\begin{table*}[htp!]
\centering
    \caption{The number of granular balls and the corresponding Accuracies by the proposed GE-GB-RVFL model under different purities.}
    \label{tab:pur2}
  \resizebox{0.85\textwidth}{!}{
\begin{tabular}{ccccccccc}
\hline
$pur$ & $1$ & $0.97$ & $0.94$ & $0.91$ & $0.88$ & $0.85$ & $0.82$ &  $0.79$ \\ \hline
{Dataset} & ACC & ACC & ACC & ACC & ACC & ACC & ACC   & ACC   \\
 & $number(GB)$ & $number(GB)$ & $number(GB)$ & $number(GB)$ & $number(GB)$ & $number(GB)$ & $number(GB)$  & $number(GB)$ \\ \hline
blood & 76 & 77.33 & 77.33 & 75.11 & 74.67  & 76.89  & 51.11 & 76.44 \\
 & 349 & 354 & 355 & 340 & 314 & 249 & 219 & 180 \\
breast\_cancer & 74.42 & 65.12 & 74.42 & 74.42 & 70.93 & 72.09 & 66.28 & 52.33 \\
 & 139 & 130 & 142 & 132 & 137 & 119 & 119 & 90 \\
breast\_cancer\_wisc\_prog & 66.67 & 58.33 & 65 & 56.67 & 63.33 & 50 & 58.33&  61.67 \\
 & 94 & 96 & 93 & 92 & 87 & 75 & 71 &  52 \\
chess\_krvkp & 92.18 & 90.82 & 90.93 & 89.47 & 93.01 & 90.09 & 90.93  & 89.57 \\
 & 1168 & 1195 & 1131 & 1062 & 969 & 858 & 757   &  619 \\
cleve & 85.56 & 78.89 & 77.78 & 67.78 & 75.56 & 74.44 & 74.44  &  70 \\
 & 108 & 113 & 114 & 111 & 107 & 68 & 43  &  37 \\ \hline
\end{tabular}}
\end{table*}

Tables \ref{tab:pur1} and \ref{tab:pur2} present the number of GBs and the corresponding accuracy achieved by GB-RVFL and GE-GB-RVFL models across different purity levels from $0.79$ to $1.0$. Notably, a clear pattern emerges: as purity decreases, the number of GBs also decreases, indicating a refinement in granularity with higher purity levels. Additionally, there's a consistent trend of increased accuracy with higher purity levels across both models.

This leads us to answer the very fundamental question in GB research: can we strike a balance between scalability and accuracy in granular computing models? Our models' scalability has already been demonstrated through experiments on NDC datasets. Furthermore, our above analyses show that our models effectively handle varying purity levels, highlighting their adaptability and effectiveness in granular computing applications. 


\begin{figure*}[ht!]
\begin{minipage}{.5\linewidth}
\centering
\subfloat[aus (GB-RVFL)]{\includegraphics[scale=0.33]{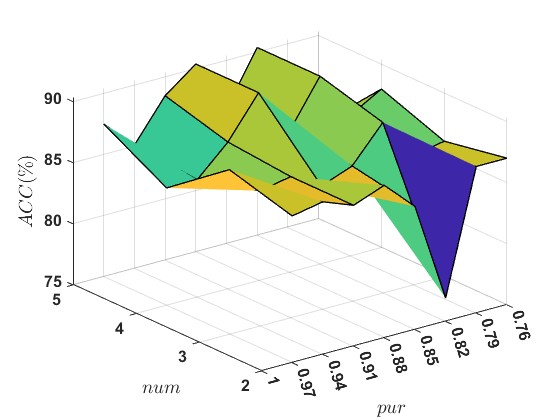}}
\end{minipage}
\begin{minipage}{.5\linewidth}
\centering
\subfloat[cleve (GB-RVFL)]{\includegraphics[scale=0.33]{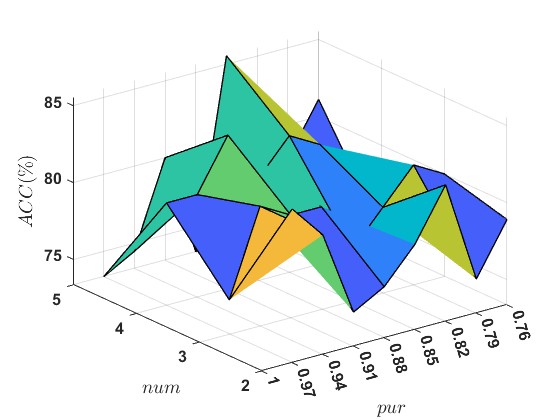}}
\end{minipage}
\par\medskip
\begin{minipage}{.5\linewidth}
\centering
\subfloat[aus (GE-GB-RVFL)]{\includegraphics[scale=0.27]{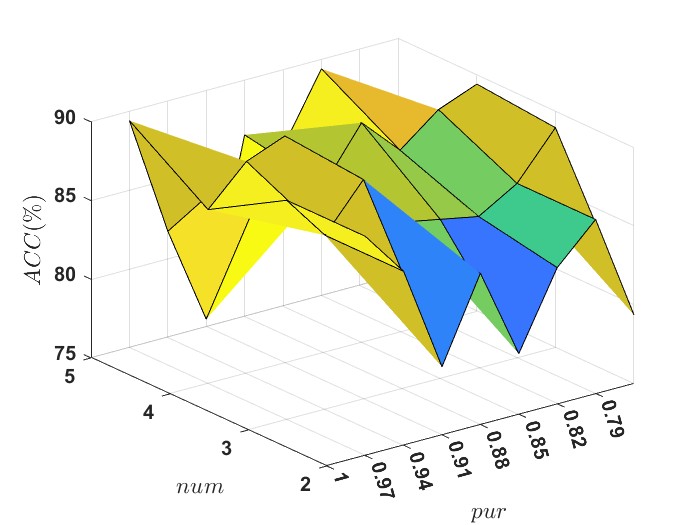}}
\end{minipage}
\begin{minipage}{.5\linewidth}
\centering
\subfloat[cleve (GE-GB-RVFL)]{\includegraphics[scale=0.33]{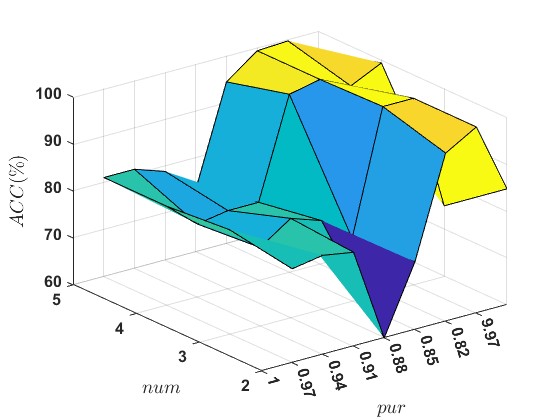}}
\end{minipage}
\caption{Effect of granular ball parameters $pur$ and $num$ on the performance of the proposed GB-RVFL and GE-GB-RVFL models.}
\label{fig:effect_pur_num}
\end{figure*}
\subsection{Effect of Granular Ball Parameters ``$pur$'' and ``$num$'' on the Performance of the Proposed GB-RVFL and GE-GB-RVFL Models}
We evaluate the impact of GB parameters $pur$ and $num$ on the performance of GB-RVFL and GE-GB-RVFL models using datasets aus and clever, as shown in Fig.  \ref{fig:effect_pur_num}.
\begin{itemize}
    \item For GB-RVFL: Examining Figs. \ref{fig:effect_pur_num}(a) and \ref{fig:effect_pur_num}(b), we note that higher values of $num$, coupled with $pur$ in the vicinity of $0.85$, yield optimal performance for the model.
    \item For GE-GB-RVFL: Analyzing Figs. \ref{fig:effect_pur_num}(c) and \ref{fig:effect_pur_num}(d), we observe that increasing $num$ values, particularly with lower $pur$, result in superior performance for the model.
\end{itemize}
These observations unveil a discernible pattern regarding the impact of $pur$ and $num$ on the proposed models. This pattern identifies a subset of parameters that, when utilized in training our model, significantly enhances its efficiency.

The remaining sensitivity analyses of hyperparameters are presented in Section S.II of the supplementary material.

\section{Conclusion and Some Potential Future Directions}
\label{conclusion}
This paper proposed the GB-RVFL model to alleviate the adverse impact of noise and outliers inherent in datasets. Further, we present the GE-GB-RVFL model, designed to uphold the intrinsic geometric structure of the dataset while preserving the fundamental properties of the GB-RVFL model. Both the proposed models, GB-RVFL and GE-GB-RVFL, take the centers of the GBs as input rather than the entire training samples. This design enhances scalability and fortifies the proposed models against the impact of noise and outliers. The efficacy of the novel GB-RVFL and GE-GB-RVFL models is showcased through extensive experimentation across multiple UCI and KEEL datasets that cover a wide range of domains and sizes. Statistical metrics such as average accuracy, sensitivity, specificity, precision and statistical tests encompassing ranking scheme, Friedman test, and Nemenyi post hoc test, consistently affirm the superior statistical performance of the GB-RVFL and GE-GB-RVFL models when compared to baseline models. We further evaluate the models over noisy environments by adding different percentages of label noise. This experiment affirms that the proposed models are robust against noise. 

Furthermore, we conducted experiments on large-scale NDC datasets. This experiment offers a comprehensive insight into the performance of the proposed GB-RVFL and GE-GB-RVFL models across a diverse spectrum of data samples, ranging from 50 thousand to 100 million. The empirical results demonstrate that the proposed GB-RVFL and GE-GB-RVFL models adaptably leverage GB to enhance the scalability and the generalization performance on large datasets. Outperforming baseline models, the proposed GB-RVFL and GE-GB-RVFL models underscore the reliability of our approach in handling large and complex datasets. Furthermore, the superior performance of the proposed models on Biomedical datasets shows their effectiveness in the real-world scenario. 

This research specifically examines shallow RVFL, which has a limited capacity for learning intrinsic feature representation within the data. Our forthcoming approach includes expanding this investigation to incorporate deep and ensemble variations of RVFL. Incorporating GE increases the complexity of the proposed GE-GB-RVFL model. Therefore, in the future, alternative methods, such as sparse GE techniques for preserving the geometrical structure of datasets, could be explored to reduce computational overhead. A potential direction for future research could involve extending the proposed models to unsupervised or semi-supervised learning settings, enabling their application in broader contexts with limited labelled data. Techniques like contrastive learning could also be integrated into the GB-RVFL framework to better exploit unlabelled data, making the model more versatile in situations with limited labelled data. Additionally, exploring alternative methods for constructing GBs could lead to further improvements in both efficiency and performance. For instance, adaptive or probabilistic approaches to GB construction could lead to better clustering while reducing computational cost. Moreover, applying these models to other challenging domains, such as time series forecasting or high-dimensional data, could unlock new opportunities and demonstrate their versatility.
\section*{Acknowledgement}
This study receives support from the Science and Engineering Research Board (SERB) through the Mathematical Research Impact-Centric Support (MATRICS) scheme Grant No. MTR/2021/000787. M. Sajid acknowledges the Council of Scientific and Industrial Research (CSIR), New Delhi, for providing fellowship under grants 09/1022(13847)/2022-EMR-I. The dataset used in this study was made possible through the Alzheimer’s Disease Neuroimaging Initiative (ADNI), funded by the Department of Defense (contract W81XWH-12-2-0012) and the National Institutes of Health (grant U01 AG024904). Support for ADNI came from the National Institute on Aging, the National Institute of Biomedical Imaging and Bioengineering, and numerous contributions from organizations:  Bristol-Myers Squibb Company; Transition Therapeutics Elan Pharmaceuticals, Inc.; IXICO Ltd.; Lundbeck;  Johnson \& Johnson Pharmaceutical Research \& Development LLC.;      Araclon Biotech; Janssen Alzheimer Immunotherapy Research \& Development, LLC.;  Novartis Pharmaceuticals Corporation; Eisai Inc.; Neurotrack Technologies; AbbVie, Alzheimer’s Association; CereSpir, Inc.; Lumosity; Biogen; Fujirebio; EuroImmun; Piramal Imaging; GE Healthcare; Cogstate; Meso Scale Diagnostics, LLC.; NeuroRx Research; Alzheimer’s Drug Discovery Foundation;  Servier; Eli Lilly and Company;  BioClinica, Inc.; Merck \& Co., Inc.; F. Hoffmann-La Roche Ltd. and its affiliated company Genentech, Inc.;   Pfizer Inc. and Takeda Pharmaceutical Company. The Canadian Institutes of Health Research supported ADNI’s clinical sites across Canada. Private sector donations were facilitated through the Foundation for the National Institutes of Health (www.fnih.org). The Northern California Institute and the Alzheimer's Therapeutic Research Institute at the University of Southern California provided support for research and teaching awards. The Neuro Imaging Laboratory at the University of Southern California facilitated the public release of the ADNI initiative's data. This investigation utilizes the ADNI dataset, which is accessible at \url{adni.loni.usc.edu}.
\bibliography{ref.bib}
\bibliographystyle{unsrtnat}

\clearpage
\section*{Supplementary Material}
\renewcommand{\thesection}{S.I}
\section{Performance Metrics} To comprehensively evaluate the effectiveness of the proposed GB-RVFL and GE-GB-RVFL models, we analyze their performance across multiple metrics, including Accuracy, Sensitivity, Specificity, and Precision.

We employ the following statistical formulas to calculate these metrics:
\begin{align}
  &\text{Accuracy}= \frac{True_+ + True_-}{True_+ + False_+ + True_- +False_-},\\
  &\text{Sensitivity}= \frac{True_+}{True_+ +False_-},\\
  &\text{Specificity}= \frac{True_-}{True_- +False_+},\\
  &\text{Precision}= \frac{True_+}{True_+ +False_+}.
\end{align}
These metrics provide insights into different aspects of model performance, with terms like false positive ($False_+$), true positive ($True_+$), false negative ($False_-$), and true negative ($True_-$) representing various outcomes in the evaluation.
\renewcommand{\thesection}{S.II}
\section{Sensitivity Analysis}
In this section, we conduct various sensitivity analyses of hyperparameters such as ``purity threshold ($pur$)'', ``activation function ($Actfun$)'', ``regularization parameter ($C$)'' and ``graph regularization parameter ($\alpha$)''.
\renewcommand{\thesubsection}{S.II.A}
\subsection{Effect of Parameter ``$pur$'' and ``$Actfun$'' on the Performance of the Proposed GB-RVFL Model}
The activation function plays a crucial role in the performance of the GB-RVFL model, particularly in its interaction with granular ball generation and its overall impact on model performance. We analyze this relationship using Fig.  \ref{Effect_pur_activation} across datasets aus, chess\_krvkp, clever, and conn\_bench\_sonar\_mines\_rocks. In these 3-D graphs, $pur$ and $Actfun$ serve as variables, with ACC representing the final output.

Our observations reveal a nuanced sensitivity to hyperparameters. For example, in the aus dataset, activation function $9$ exhibits the lowest performance within the purity range of $0.88$ to $0.91$. Conversely, in the chess\_krvkp dataset, activation function $9$ demonstrates superior performance within the purity range of $0.85$ to $0.88$. Thus, we observe the mix performance with respect to $Actfun$, and no particular patterns have been found. These findings highlight the need for fine-tuning the activation function to achieve better results.
\renewcommand{\thefigure}{S.1}
\begin{figure*}[ht!]
\begin{minipage}{.5\linewidth}
\centering
\subfloat[aus]{\includegraphics[scale=0.3]{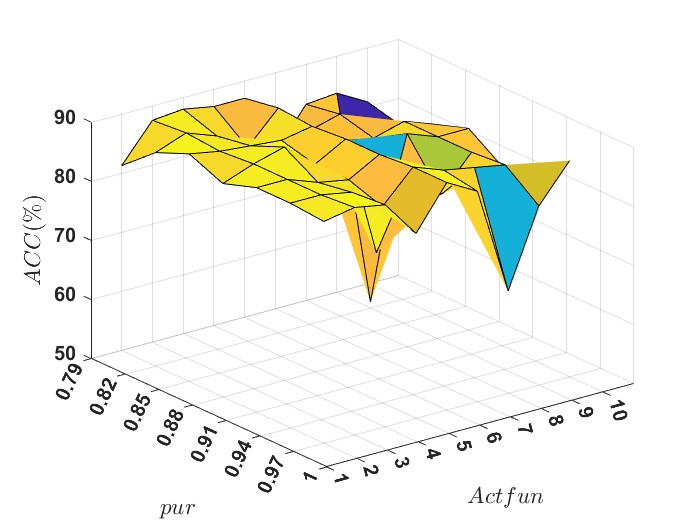}}
\end{minipage}
\begin{minipage}{.5\linewidth}
\centering
\subfloat[chess\_krvkp]{\includegraphics[scale=0.3]{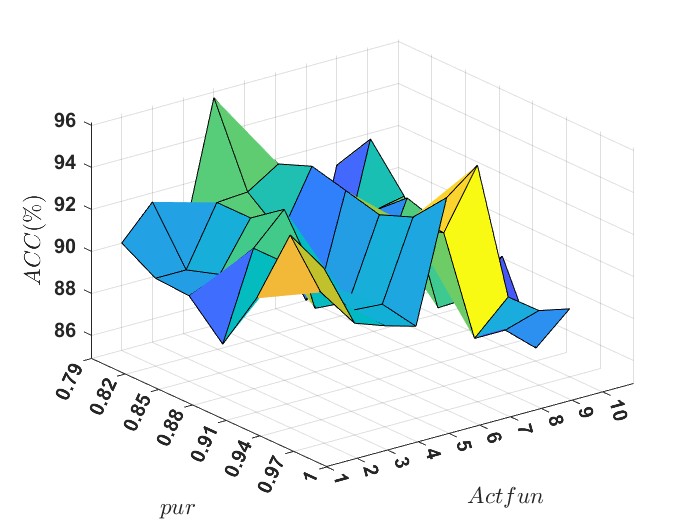}}
\end{minipage}
\par\medskip
\begin{minipage}{.5\linewidth}
\centering
\subfloat[cleve]{\includegraphics[scale=0.3]{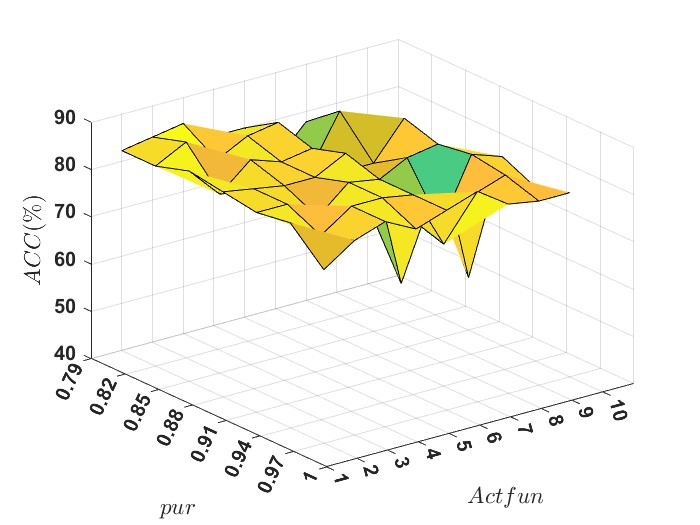}}
\end{minipage}
\begin{minipage}{.5\linewidth}
\centering
\subfloat[conn\_bench\_sonar\_mines\_rocks]{\includegraphics[scale=0.3]{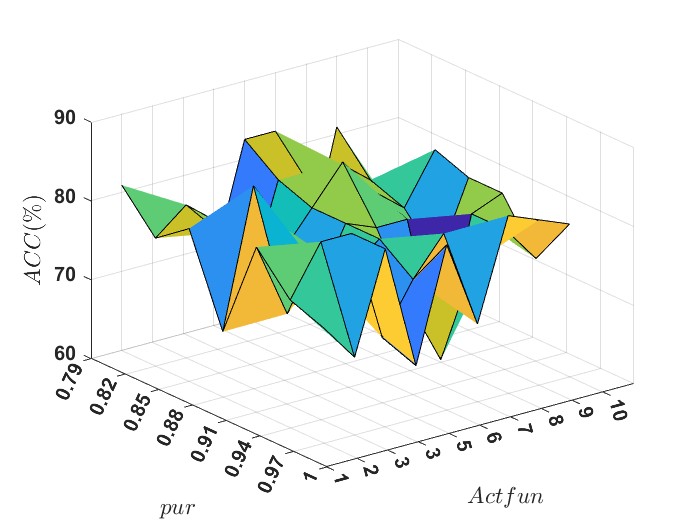}}
\end{minipage}
\caption{Effect of parameters purity $(pur)$ and activation function $(Act~fun)$ on the performance of the proposed GB-RVFL model.}
\label{Effect_pur_activation}
\end{figure*}
\renewcommand{\thefigure}{S.2}
\begin{figure*}[ht!]
\begin{minipage}{.5\linewidth}
\centering
\subfloat[aus (GB-RVFL)]{\includegraphics[scale=0.35]{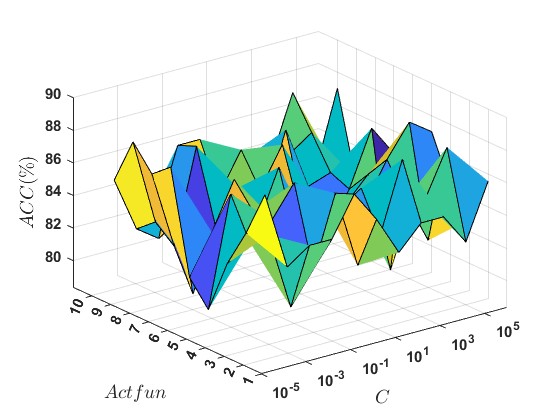}}
\end{minipage}
\begin{minipage}{.5\linewidth}
\centering
\subfloat[bank (GB-RVFL)]{\includegraphics[scale=0.35]{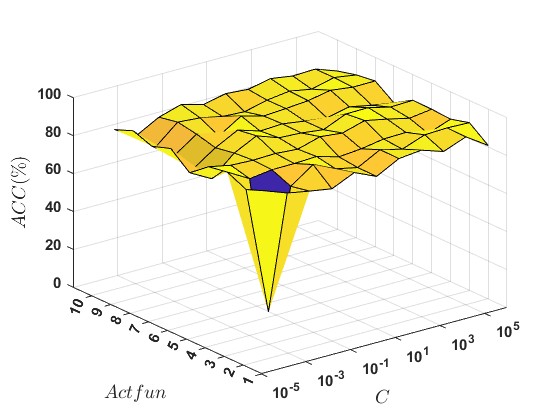}}
\end{minipage}
\par\medskip
\begin{minipage}{.5\linewidth}
\centering
\subfloat[aus (GE-GB-RVFL)]{\includegraphics[scale=0.3]{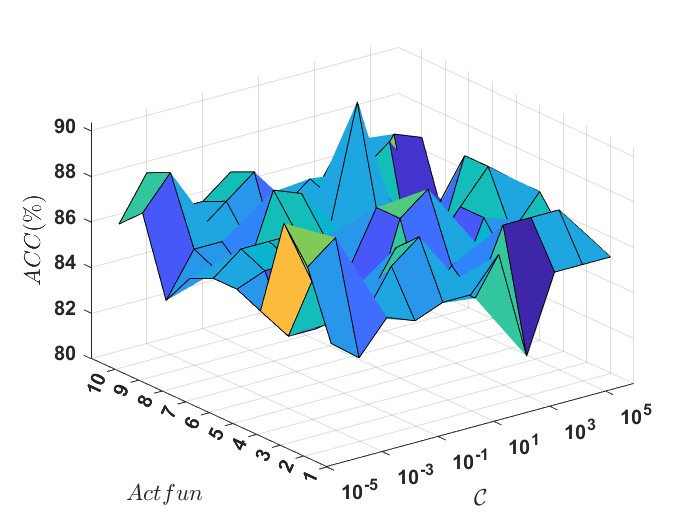}}
\end{minipage}
\begin{minipage}{.5\linewidth}
\centering
\subfloat[bank (GE-GB-RVFL)]{\includegraphics[scale=0.3]{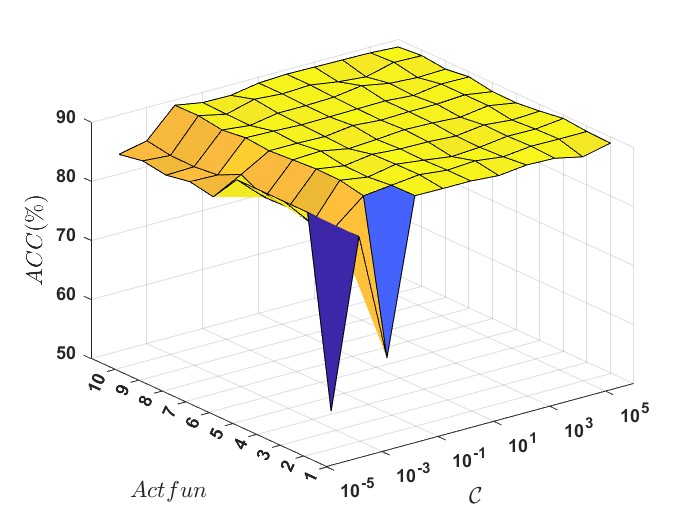}}
\end{minipage}
\caption{Effect of parameters $\mathcal{C}$ and activation function $(Act~fun)$ on the performance of the proposed GB-RVFL and GE-GB-RVFL models.}
\label{effect_c_act}
\end{figure*}
\renewcommand{\thefigure}{S.3}
\begin{figure*}[ht!]
\begin{minipage}{.5\linewidth}
\centering
\subfloat[ecoli2]{\includegraphics[scale=0.4]{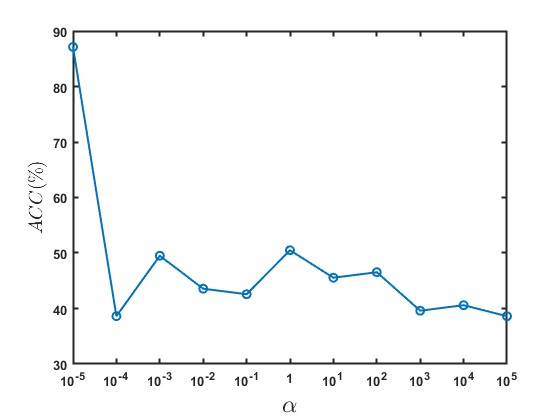}}
\end{minipage}
\begin{minipage}{.5\linewidth}
\centering
\subfloat[monks\_3]{\includegraphics[scale=0.4]{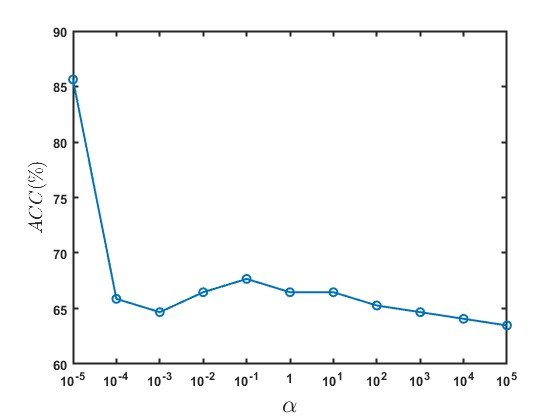}}
\end{minipage}
\par\medskip
\begin{minipage}{.5\linewidth}
\centering
\subfloat[new-thyroid1]{\includegraphics[scale=0.40]{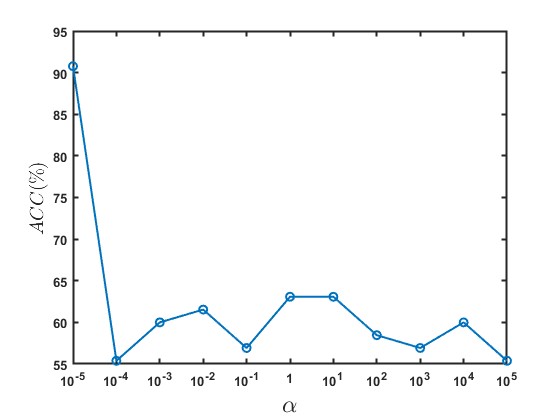}}
\end{minipage}
\begin{minipage}{.5\linewidth}
\centering
\subfloat[tic\_tac\_toe]{\includegraphics[scale=0.40]{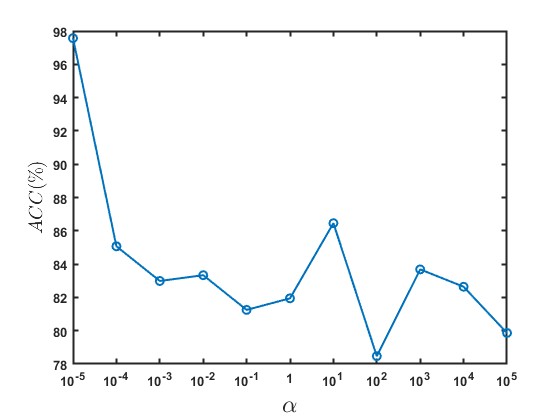}}
\end{minipage}
\caption{Effect of parameters $\alpha$ on the performance of the proposed GE-GB-RVFL model.}
\label{effect_alpha}
\end{figure*}
\renewcommand{\thesubsection}{S.II.B}
\subsection{Effect of Parameters ``$\mathcal{C}$'' and ``$Actfun$'' on the performance of the proposed GB-RVFL and GE-GB-RVFL models}
To evaluate the impact of the regularization parameter $\mathcal{C}$ on the performance of GB-RVFL and GE-GB-RVFL models, we present graphs in Fig.  \ref{effect_c_act} with $\mathcal{C}$ and $Actfun$ as independent variables and ACC as the dependent variable across the aus and bank datasets. Our analysis reveals interesting insights. In the bank datasets, both GB-RVFL and GE-GB-RVFL models demonstrate consistent performance, except for instances where $\mathcal{C}$ equals $10^{-3}$ and/or $10^{-1}$, where some variability is observed. However, on the aus dataset, the performance of the proposed models shows sensitivity to hyperparameters, indicating a dataset-dependent behavior. This underscores the importance of tuning the parameter $\mathcal{C}$ to achieve optimal results based on the specific dataset characteristics. In conclusion, our results underscore the impact of dataset characteristics on the performance of the proposed models, stressing the importance of fine-tuning the regularization parameter $\mathcal{C}$ for optimal model performance.
\renewcommand{\thesubsection}{S.II.C}
\subsection{Effect of Parameters ``$\alpha$'' on the performance of the proposed GE-GB-RVFL model}
The effect of the hyperparameter $\alpha$ on the GE-GB-RVFL model's performance is depicted in Fig. \ref{effect_alpha}. Our analysis demonstrates a clear pattern: as the lowest value of $\alpha$, there is a significant improvement in the model's performance, particularly when $\alpha = 10^{-5}$. Beyond this point, however, the performance tends to level off, as indicated by the accuracy (ACC) metrics. This plateau suggests diminishing returns in performance gains for larger values of $\alpha$.

The results emphasize that the optimal performance of the GE-GB-RVFL model is highly dependent on the careful selection and fine-tuning of $\alpha$. Small values of $\alpha$ contribute positively to performance, but further increases beyond a certain threshold offer minimal additional benefit. This sensitivity of the model to $\alpha$ highlights the importance of considering the characteristics of the dataset when tuning the model. Different datasets may respond differently to variations in $\alpha$, meaning that optimal performance may vary depending on the specific features and patterns present in the data. Thus, fine-tuning $\alpha$ is crucial for ensuring that the GE-GB-RVFL model reaches its full potential in terms of accuracy and robustness across diverse datasets.
\renewcommand{\thetable}{S.1}
\begin{table*}[ht!]
\centering
\caption{Index of activation functions.}
\label{tab:activation_function}
\resizebox{0.85\linewidth}{!}{
\begin{tabular}{|c|l|}
\hline
\textbf{Index} & \multicolumn{1}{c|}{\textbf{Activation Functions}}    \\ \hline
1              & Scaled Exponential Linear Unit (SELU)                 \\ \hline
2              & Rectified Linear Units (ReLU)                         \\ \hline
3              & Sigmoid                                               \\ \hline
4              & Sine (Sin)                                            \\ \hline
5              & Hard Limit Transfer Function (Hardlim)                \\ \hline
6              & Triangular Basis Transfer Function (Tribas)           \\ \hline
7              & Radial Basis Transfer Function (Radbas)               \\ \hline
8              & Signum (Sgn)                                          \\ \hline
9              & Leaky Rectified Linear Unit, or (Leaky ReLU) \\ \hline
10              & Hyperbolic Tangent Sigmoid Transfer Function (Tansig)  \\ \hline
\end{tabular}}
\end{table*}
\renewcommand{\thetable}{S.2}
\begin{landscape}
\begin{table*}[htp!]
\centering\caption{Best hyperparameters for all the compared models for the experiments on UCI and KEEL datasets.}
    \label{UCI and KEEL Hypermeter}
\resizebox{1.1\linewidth}{!}{
\begin{tabular}{l|cccccccc}
\hline
Dataset & RVFL \cite{pao1994learning} & RVFLwoDL \cite{huang2006extreme} & IF-RVFL \cite{malik2022alzheimer} & NF-RVFL \cite{sajid2024neuro} & Wave-RVFL \cite{sajid2024wavervfl} & GB-RVFLwoDL$^{\bigstar}$                 & GB-RVFL$^{\bigstar}$                 & GE-GB-RVFL$^{\bigstar}$ \\
 & $(\mathcal{C}, N, act fun)$ & $(\mathcal{C}, N, act fun)$ & $(\mathcal{C}, N, \mu, act fun)$ & $(\mathcal{C}, N, J, act fun)$ & $(\mathcal{C}, N, act fun, a, b, \delta)$ & $(\mathcal{C}, N, act fun)$ & $(\mathcal{C}, N, act fun)$ & $(\mathcal{C}, \alpha, N, act fun)$ \\ \hline
acute\_inflammation & $(0.00001, 3, 3)$ & $(0.00001, 83, 8)$ & $(10, 103, 2, 2)$ & $(0.00001, 3, 10, 4)$ & $(1, 103, 5, 0, 1.6, 1)$ & $(1000, 3, 4)$ & $(0.01, 143, 8)$ & $(0.00001,   1, 203, 4)$ \\
aus & $(0.001, 23, 1)$ & $(0.001, 23, 1)$ & $(100, 103, 2, 4)$ & $(0.01, 3, 5, 8)$ & $(1, 23, 5, -0.5, 1.6, 1)$ & $(0.01, 183, 9)$ & $(0.01, 203, 8)$ & $(0.01,   0.01, 23, 1)$ \\
bank & $(1, 163, 1)$ & $(1000, 143, 1)$ & $(100, 23, 16, 4)$ & $(10000, 43, 35, 7)$ & $(100, 203, 5, 0, 1.6, 1)$ & $(1, 163, 3)$ & $(100000, 23, 1)$ & $(1,   0.0001, 23, 8)$ \\
chess\_krvkp & $(100000, 183, 2)$ & $(0.001, 183, 1)$ & $(100000, 203, 8, 9)$ & $(10000, 83, 15, 2)$ & $(0.0001, 183, 6, 2.5, 0.6, 1)$ & $(0.1, 103, 1)$ & $(0.01, 183, 9)$ & $(0.01,   0.0001, 163, 1)$ \\
cleve & $(0.00001, 203, 8)$ & $(0.00001, 203, 8)$ & $(1000, 43, 0.5, 8)$ & $(0.01, 183, 5, 4)$ & $(10000, 83, 2, 0, 0.1, 1)$ & $(0.00001, 103, 8)$ & $(100, 203, 7)$ & $(0.00001,   0.001, 63, 1)$ \\
conn\_bench\_sonar\_mines\_rocks & $(1000, 203, 1)$ & $(1000, 203, 1)$ & $(100000, 123, 0.03125, 2)$ & $(0.1, 143, 35, 6)$ & $(0.01, 3, 3, -0.5, 0.85, 1)$ & $(1, 63, 9)$ & $(1, 63, 9)$ & $(10000,   0.001, 43, 9)$ \\
crossplane130 & $(0.1, 43, 8)$ & $(0.1, 43, 8)$ & $(1000, 43, 0.125, 6)$ & $(0.1, 23, 20, 7)$ & $(1, 23, 5, -0.5, 1.6, 1)$ & $(0.1, 103, 8)$ & $(0.1, 103, 8)$ & $(0.00001,   1, 3, 8)$ \\
echocardiogram & $(0.001, 183, 2)$ & $(0.01, 203, 3)$ & $(0.1, 3, 0.125, 8)$ & $(1000, 203, 25, 5)$ & $(100, 123, 6, -0.5, 0.35, 1)$ & $(0.1, 23, 2)$ & $(0.1, 3, 2)$ & $(0.00001,   0.1, 23, 6)$ \\
ecoli0137vs26 & $(0.0001, 83, 9)$ & $(0.01, 63, 9)$ & $(10000, 163, 2, 2)$ & $(1, 203, 35, 2)$ & $(1, 103, 1, 0.5, 1.6, 1)$ & $(1000, 3, 6)$ & $(10, 3, 1)$ & $(0.00001,   10, 43, 2)$ \\
ecoli-0-1-4-6\_vs\_5 & $(0.01, 103, 4)$ & $(0.01, 103, 4)$ & $(10000, 123, 0.5, 9)$ & $(10, 163, 30, 4)$ & $(0.1, 23, 1, 1, 1.1, 1)$ & $(0.1, 123, 7)$ & $(0.1, 123, 7)$ & $(0.1,   1, 103, 7)$ \\
ecoli2 & $(0.01, 43, 4)$ & $(0.01, 43, 4)$ & $(10, 43, 4, 7)$ & $(0.01, 183, 15, 9)$ & $(0.01, 83, 2, -2, 1.6, 0.0001)$ & $(0.1, 83, 9)$ & $(0.1, 63, 5)$ & $(0.1,   0.00001, 163, 5)$ \\
fertility & $(0.01, 63, 1)$ & $(1000, 203, 7)$ & $(100000, 103, 0.25, 4)$ & $(100000, 123, 45, 3)$  & $(1, 203, 3, -0.5, 1.6, 1)$ & $(0.00001, 3, 2)$ & $(0.00001, 3, 3)$ & $(0.001,   10000, 23, 4)$ \\
haberman\_survival & $(0.0001, 143, 9)$ & $(10, 3, 3)$ & $(10, 143, 8, 9)$ & $(1000, 143, 40, 4)$ & $(1, 83, 2, 3, 1.1, 1)$ & $(0.1, 23, 1)$ & $(0.1, 23, 1)$ & $(10000,   100, 3, 3)$ \\
heart\_hungarian & $(0.00001, 123, 4)$ & $(0.01, 143, 2)$ & $(10, 3, 2, 4)$ & $(0.001, 183, 5, 4)$ & $(1, 183, 6, -2, 0.85, 1)$ & $(0.00001, 23, 8)$ & $(0.00001, 3, 8)$ & $(0.00001,   0.01, 163, 4)$ \\
heart-stat & $(0.1, 23, 1)$ & $(0.01, 163, 5)$ & $(0.0001, 3, 0.0625, 7)$ & $(10000, 3, 20, 3)$  & $(0.00001, 3, 2, 3, 1.1, 1)$ & $(0.001, 83, 4)$ & $(0.001, 83, 4)$ & $(0.01,   1, 103, 2)$ \\
ionosphere & $(0.01, 163, 9)$ & $(100, 43, 2)$ & $(10000, 83, 16, 5)$ & $(0.1, 183, 15, 3)$  & $(1000, 103, 1, -2, 0.1, 1)$ & $(1, 23, 7)$ & $(10000, 43, 9)$ & $(0.1,   10, 123, 2)$ \\
led7digit-0-2-4-5-6-7-8-9\_vs\_1 & $(1, 23, 5)$ & $(0.01, 83, 1)$ & $(0.1, 63, 8, 3)$ & $(0.1, 123, 40, 4)$  & $(0.01, 23, 3.5, 3.5, 0.1, 1)$ & $(0.1, 83, 2)$ & $(1, 123, 6)$ & $(0.1,   0.01, 63, 7)$ \\
mammographic & $(0.1, 143, 4)$ & $(0.1, 143, 4)$ & $(10, 63, 16, 8)$ & $(1000, 203, 20, 4)$  & $(0.001, 203, 3, 0.5, 0.35, 1)$ & $(10, 63, 1)$ & $(10, 3, 9)$ & $(1000,   0.001, 3, 9)$ \\
monk1 & $(100000, 3, 8)$ & $(1000, 3, 6)$ & $(100, 63, 4, 2)$ & $(100, 3, 10, 2)$ & $(0.01, 103, 1, 0.5, 0.85, 1)$ & $(100000, 183, 2)$ & $(10, 123, 9)$ & $(1000,   0.00001, 23, 5)$ \\
monk3 & $(0.00001, 43, 1)$ & $(0.00001, 103, 1)$ & $(1, 3, 0.25, 8)$ & $(0.00001, 143, 50, 7)$ &  $(10000, 103, 2, 3.5, 0.85, 1)$ & $(10, 203, 1)$ & $(10, 203, 1)$ & $(0.00001,   0.0001, 163, 8)$ \\
new-thyroid1 & $(10, 163, 1)$ & $(10, 163, 1)$ & $(100, 143, 0.5, 2)$ & $(0.1, 123, 15, 5)$ & $(0.01, 63, 3, 2.5, 1.1, 1)$ & $(0.1, 163, 9)$ & $(100, 23, 7)$ & $(0.1,   0.00001, 3, 1)$ \\
pima & $(0.0001, 183, 9)$ & $(0.0001, 183, 9)$ & $(10000, 163, 2, 8)$ & $(1000, 23, 40, 4)$ & $(0.01, 163, 5, 3, 0.35, 1)$ & $(0.01, 23, 2)$ & $(0.1, 43, 6)$ & $(100000,   0.00001, 3, 8)$ \\
statlog\_heart & $(0.1, 23, 1)$ & $(0.01, 163, 5)$ & $(0.0001, 3, 0.25, 7)$ & $(10000, 3, 20, 6)$ & $(0.0001, 83, 2, -2, 1.6, 1)$ & $(0.001, 83, 4)$ & $(0.001, 83, 4)$ & $(0.01,   1, 103, 2)$ \\
tic\_tac\_toe & $(100, 183, 2)$ & $(1000, 183, 9)$ & $(100000, 63, 2, 2)$ & $(0.1, 103, 15, 2)$ & $(0.00001, 203, 1, 0.5, 1.85, 0.01)$ & $(1000, 183, 2)$ & $(1000, 183, 2)$ & $(0.01,   1000, 143, 9)$ \\
transfusion & $(0.1, 123, 5)$ & $(1, 43, 9)$ & $(10000, 43, 0.5, 6)$ & $(0.01, 163, 50, 3)$ & $(1000, 183, 6, 2.5, 0.6, 1)$ & $(0.001, 183, 9)$ & $(0.1, 23, 8)$ & $(0.001,   0.01, 143, 9)$ \\
vehicle2 & $(100000, 143, 4)$ & $(1, 183, 3)$ & $(10000, 163, 0.125, 4)$ & $(100000, 103, 25, 2)$ & $(1000, 203, 5, -0.5, 1.6, 1)$ & $(0.1, 143, 9)$ & $(0.1, 143, 9)$ & $(0.1,   0.00001, 183, 1)$ \\
vertebral\_column\_2clases & $(100, 3, 4)$ & $(10, 23, 9)$ & $(1000, 183, 2, 8)$ & $(1, 43, 45, 5)$ & $(0.01, 203, 6, 5, 0.85, 1)$ & $(100, 3, 4)$ & $(1, 143, 4)$ & $(1,   0.01, 3, 6)$ \\
wpbc & $(0.1, 103, 5)$ & $(0.1, 143, 3)$ & $(100000, 143, 16, 6)$ & $(10, 203, 25, 9)$ & $(0.1, 23, 1, -2, 1.1, 1)$ & $(10, 63, 4)$ & $(0.001, 3, 2)$ & $(0.1,   100, 3, 3)$ \\
yeast-0-2-5-7-9\_vs\_3-6-8 & $(1, 83, 7)$ & $(1, 123, 1)$ & $(10000, 63, 16, 5)$ & $(100000, 63, 45, 2)$ & $(1, 23, 5, -0.5, 1.6, 1)$ & $(0.1, 63, 3)$ & $(10000, 3, 3)$ & $(1000,   1, 23, 4)$ \\
yeast1 & $(0.1, 83, 1)$ & $(0.00001, 3, 1)$ & $(10, 123, 32, 3)$ & $(0.01, 203, 20, 5)$ & $(0.001, 3, 2, 1, 1.85, 1)$ & $(0.1, 43, 2)$ & $(1, 83, 1)$ & $(0.01,   0.00001, 203, 9)$ \\ \hline
\multicolumn{6}{l}{$^{\bigstar}$ denotes the proposed models.}\\
\end{tabular}}
\end{table*}
\end{landscape}
\renewcommand{\thetable}{S.3}
\begin{landscape}
\begin{table*}[htp!]
\centering
    \caption{Best hyperparameters for all the compared models for the experiments on UCI and KEEL datasets with noise.}
    \label{UCI and KEEL Hypermeter with noise}
  \resizebox{1.1\linewidth}{!}{
\begin{tabular}{l|c|cccccc}
\hline
Dataset & Noise & RVFL \cite{pao1994learning} & RVFLwoDL \cite{huang2006extreme} & Wave-RVFL \cite{sajid2024wavervfl}  & GB-RVFLwoDL$^{\bigstar}$                 & GB-RVFL$^{\bigstar}$                 & GE-GB-RVFL$^{\bigstar}$ \\
 &  & $(\mathcal{C}, N, act fun)$ & $(\mathcal{C}, N, act fun)$ & $(\mathcal{C}, N, act fun)$ & $(\mathcal{C}, N, act fun, a, b, \delta)$ & $(\mathcal{C}, N, act fun)$ & $(\mathcal{C}, \alpha, N, act fun)$ \\ \hline
conn\_bench\_sonar\_mines\_rocks & $5\%$ & $(0.1, 43, 1)$ & $(10, 183, 2)$ & $(1, 23, 3, 0, 1.35, 1)$ & $(1000, 183, 9)$ & $(1000, 183, 9)$ & $(10000, 100, 183, 4)$ \\
 & $10\%$ & $(0.1, 183, 8)$ & $(0.001, 183, 2)$ & $(1, 63, 4, 0.5, 1.85, 1)$ & $(10000, 43, 1)$ & $(10000, 43, 1)$ & $(1, 100, 163, 4)$ \\
 & $20\%$ & $(0.0001, 183, 8)$ & $(0.0001, 183, 8)$ & $(0.00001, 23, 5, 0, 0.6, 1)$ & $(0.1, 43, 8)$ & $(0.001, 123, 1)$ & $(0.0001, 1000, 103, 8)$ \\
 & $30\%$ & $(0.01, 183, 2)$ & $(0.01, 183, 2)$ & $(0.1, 83, 5, -0.5, 0.1, 1)$ & $(0.01, 83, 8)$ & $(0.1, 103, 5)$ & $(0.01, 0.01, 103, 8)$ \\
 & $40\%$ & $(0.001, 143, 4)$ & $(100000, 43, 4)$ & $(0.001, 23, 5, -2, 0.85, 0.001)$ & $(1000, 123, 2)$ & $(0.1, 183, 4)$ & $(0.0001, 100, 63, 4)$ \\ \hline
ecoli-0-1-4-6\_vs\_5 & $5\%$ & $(0.1, 183, 9)$ & $(0.1, 183, 9)$ & $(100, 23, 4, 0, 0.1, 1)$ & $(0.0001, 3, 9)$ & $(0.0001, 43, 9)$ & $(0.00001, 0.1, 63, 2)$ \\
 & $10\%$ & $(0.01, 23, 4)$ & $(0.01, 43, 4)$ & $(0.001, 3, 3, 0, 0.1, 0.0001)$ & $(0.1, 63, 2)$ & $(0.1, 63, 6)$ & $(0.00001, 10, 143, 6)$ \\
 & $20\%$ & $(0.01, 123, 9)$ & $(0.01, 123, 9)$ & $(10, 3, 1, 0.5, 0.6, 0.001)$ & $(0.001, 203, 9)$ & $(0.001, 203, 9)$ & $(0.0001, 0.01, 3, 9)$ \\
 & $30\%$ & $(0.0001, 3, 7)$ & $(0.0001, 3, 8)$ & $(10, 3, 1, 0.5, 0.35, 0.01)$ & $(1000, 103, 6)$ & $(0.0001, 3, 5)$ & $(100, 0.01, 3, 2)$ \\
 & $40\%$ & $(0.001, 43, 3)$ & $(0.01, 23, 6)$ & $(1, 123, 5, 1, 1.35, 1)$ & $(100, 123, 4)$ & $(1, 23, 7)$ & $(10000, 0.00001, 23, 3)$ \\ \hline
heart-stat & $5\%$ & $(0.01, 203, 5)$ & $(0.001, 143, 8)$ & $(0.001, 23, 5, -1, 0.85, 0.0001)$ & $(0.0001, 63, 4)$ & $(10, 23, 8)$ & $(10000, 10000, 143, 4)$ \\
 & $10\%$ & $(0.001, 63, 1)$ & $(0.001, 63, 1)$ & $(0.0001, 23, 5, -1, 1.85, 1)$ & $(0.00001, 103, 4)$ & $(0.0001, 143, 8)$ & $(0.001, 0.00001, 3, 1)$ \\
 & $20\%$ & $(100, 23, 1)$ & $(100, 23, 1)$ & $(0.01, 23, 5, -1, 1.85, 0.0001)$ & $(0.00001, 183, 8)$ & $(0.001, 123, 8)$ & $(100000, 0.01, 183, 8)$ \\
 & $30\%$ & $(0.00001, 3, 1)$ & $(0.00001, 43, 8)$ & $(10, 63, 2, -0.5, 1.1, 0.0001)$ & $(0.001, 103, 4)$ & $(0.0001, 123, 4)$ & $(0.00001, 0.0001, 123, 4)$ \\
 & $40\%$ & $(100000, 63, 4)$ & $(100000, 163, 5)$ & $(10000, 203, 2, 0, 0.1, 1)$ & $(0.001, 63, 4)$ & $(0.00001, 3, 4)$ & $(10, 0.01, 203, 4)$ \\ \hline
yeast-0-2-5-7-9\_vs\_3-6-8 & $5\%$ & $(0.01, 83, 4)$ & $(100, 23, 9)$ & $(1, 23, 5, 0.5, 1.6, 1)$ & $(1, 83, 1)$ & $(100, 3, 9)$ & $(1, 100, 203, 3)$ \\
 & $10\%$ & $(0.01, 23, 2)$ & $(1, 43, 1)$ & $(1000, 3, 2, 0, 1.6, 1)$ & $(100, 23, 6)$ & $(0.01, 3, 2)$ & $(1, 1, 63, 3)$ \\
 & $20\%$ & $(1, 43, 7)$ & $(0.1, 163, 3)$ & $(0.01, 103, 1, 0.5, 1.85, 1)$ & $(0.1, 63, 1)$ & $(0.1, 43, 4)$ & $(0.1, 0.0001, 63, 5)$ \\
 & $30\%$ & $(0.01, 43, 2)$ & $(0.01, 43, 2)$  & $(100, 103, -2, 0.1, 1.85, 1)$ & $(1000, 3, 2)$ & $(0.00001, 3, 3)$ & $(0.1, 0.01, 183, 5)$ \\
 & $40\%$ & $(0.001, 103, 9)$ & $(0.001, 103, 9)$  & $(0.0001, 203, 5, 3, 0.85, 0.01)$ & $(0.1, 23, 5)$ & $(10000, 3, 1)$ & $(100, 0.0001, 3, 4)$ \\ \hline
\multicolumn{6}{l}{$^{\bigstar}$ denotes the proposed models.}\\
\end{tabular}}
\end{table*}
\end{landscape}
\renewcommand{\thetable}{S.4}
\begin{landscape}
\begin{table*}
\centering
\caption{Best hyperparameters for all the compared models for the experiments on NDC datasets.}
\label{tab:NDC_Accuracy}
\resizebox{1.1\textwidth}{!}{
\begin{tabular}{l|ccccc}\hline \\ 
Dataset & RVFL \cite{pao1994learning} & RVFLwoDL \cite{huang2006extreme} & GB-RVFLwoDL$^{\bigstar}$                 & GB-RVFL$^{\bigstar}$                 & GE-GB-RVFL$^{\bigstar}$ \\
 & $(\mathcal{C}, N, act fun)$ & $(\mathcal{C}, N, act fun)$ & $(\mathcal{C}, N, act fun)$ & $(\mathcal{C}, N, act fun)$ & $(\mathcal{C}, \alpha, N, act fun)$ \\ \hline
NDC-50K & $(0.1, 203, 9)$ & $(0.1, 203, 9)$ & $(0.1, 183, 2)$ & $(0.1, 123, 9)$ & $(0.1, 0.00001, 163, 9)$ \\
NDC-100K & $(100000, 203, 2)$ & $(0.01, 203, 2)$ & $(0.1, 203, 9)$ & $(0.1, 203, 9)$ & $(1, 0.01, 123, 9)$ \\
NDC-500K & $(100, 203, 9)$ & $(100, 203, 9)$ & $(0.1, 163, 1)$ & $(10, 183, 2)$ & $(0.1, 0.00001, 203, 2)$ \\
NDC-1M & $(0.01, 203, 9)$ & $(0.01, 203, 9)$ & $(100000, 163, 9)$ & $(10, 183, 2)$ & $(100, 10, 203, 9)$ \\
NDC-3M & $(0.001, 203, 2)$ & $(0.01, 203, 9)$ & $(10, 203, 2)$ & $(10, 203, 2)$ & $(1, 1, 203, 9)$ \\
NDC-5M & $(0.001, 203, 2)$ & $(0.00001, 203, 9)$ & $(100, 183, 9)$ & $(100, 183, 9)$ & $(100, 1, 183, 2)$ \\
NDC-10M & $(1000, 203, 9)$ & $(0.1, 203, 9)$ & $(100000, 203, 2)$ & $(1000, 203, 9)$ & $(100, 1, 183, 2)$ \\
NDC-30M & - & - & $(1, 203, 2)$ & $(1, 203, 2)$ & $(10, 1000, 183, 9)$ \\
NDC-50M & -  & - & $(100, 203, 1)$ & $(10000, 203, 2)$ & $(1000, 0.001, 203, 9)$ \\
NDC-100M & - & - & $(10000, 203, 1)$ & $(0.1, 203, 2)$ & $(1000, 0.1, 203, 9)$ \\ \hline
\multicolumn{6}{l}{$^{\bigstar}$ denotes the proposed models.}\\
\end{tabular}%
}
\end{table*}
\end{landscape}
\renewcommand{\thetable}{S.5}
\begin{landscape}
\begin{table*}
\centering
\caption{Best hyperparameters for all the compared models for the experiments on biomedical datasets.}
\label{tab:Biomedical_Accuracy}
\resizebox{1.1\linewidth}{!}{
\begin{tabular}{lcccccc}
\hline
Dataset & RVFL \cite{pao1994learning} & RVFLwoDL \cite{huang2006extreme} & IF-RVFL \cite{malik2022alzheimer} & GB-RVFLwoDL$^{\bigstar}$                 & GB-RVFL$^{\bigstar}$                 & GE-GB-RVFL$^{\bigstar}$ \\
 & $(\mathcal{C}, N, act fun)$ & $(\mathcal{C}, N, act fun)$ & $(\mathcal{C}, N, \mu, act fun)$ &   $(\mathcal{C}, N, act fun)$ & $(\mathcal{C}, N, act fun)$ & $(\mathcal{C}, \alpha, N, act fun)$ \\ \hline
adenosis\_vs\_ductal\_carcinoma & $(0.0001, 43, 5)$ & $(0.001, 83, 8)$ & $(0.01, 123, 0.03125, 9)$ &   $(0.00001, 163, 8)$ & $(0.00001, 3, 4)$ & $(0.001, 0.01, 143, 1)$ \\
adenosis\_vs\_lobular\_carcinoma & $(100000, 163, 1)$ & $(100000, 163, 1)$ & $(0.1, 23, 4, 2)$ &   $(100000, 43, 2)$ & $(0.0001, 163, 2)$ & $(0.1, 0.0001, 83, 1)$ \\
adenosis\_vs\_mucinous\_carcinoma & $(0.00001, 103, 1)$ & $(10, 3, 8)$ & $(0.01, 163, 0.0625, 8)$ &   $(1, 3, 4)$ & $(10, 163, 1)$ & $(0.00001, 0.0001, 23, 1)$ \\
adenosis\_vs\_papillary\_carcinoma & $(0.0001, 123, 2)$ & $(100000, 43, 6)$ & $(10, 3, 2, 5)$ &   $(0.0001, 103, 4)$ & $(100000, 123, 9)$ & $(0.001, 10, 23, 1)$ \\
CN\_vs\_AD & $(0.01, 23, 8)$ & $(0.001, 103, 1)$ & $(0.01, 103, 8, 2)$ &   $(0.1, 163, 3)$ & $(1000, 183, 7)$ & $(0.00001, 100, 183, 6)$ \\
CN\_vs\_MCI & $(0.0001, 163, 9)$ & $(0.0001, 163, 9)$ & $(0.001, 203, 16, 2)$ &   $(0.001, 103, 1)$ & $(0.001, 83, 1)$ & $(0.001, 0.0001, 203, 9)$ \\
fibroadenoma\_vs\_ductal\_carcinoma & $(0.0001, 23, 8)$ & $(0.0001, 143, 8)$ & $(0.001, 143, 2, 7)$ &   $(0.001, 203, 2)$ & $(0.00001, 3, 3)$ & $(0.00001, 1, 163, 4)$ \\
fibroadenoma\_vs\_lobular\_carcinoma & $(0.00001, 103, 2)$ & $(10, 3, 7)$ & $(100, 123, 0.5, 5)$   & $(0.0001, 103, 4)$ & $(0.01, 183, 1)$ & $(0.1, 10, 123, 1)$ \\
fibroadenoma\_vs\_mucinous\_carcinoma & $(100000, 143, 9)$ & $(1000, 43, 3)$ & $(0.01, 103, 32, 5)$   & $(100000, 23, 7)$ & $(0.00001, 183, 8)$ & $(0.0001, 0.00001, 3, 9)$ \\
fibroadenoma\_vs\_papillary\_carcinoma & $(0.00001, 3, 9)$ & $(100000, 43, 9)$ & $(0.001, 203, 32, 2)$   & $(0.01, 203, 1)$ & $(0.001, 123, 2)$ & $(0.00001, 1, 43, 1)$ \\
MCI\_vs\_AD & $(0.001, 103, 9)$ & $(0.001, 103, 9)$ & $(0.001, 103, 32, 2)$   & $(1, 203, 3)$ & $(10000, 23, 3)$ & $(0.1, 0.0001, 23, 8)$ \\
phyllodes\_tumour\_vs\_ductal\_carcinoma & $(0.0001, 63, 3)$ & $(0.00001, 203, 8)$ & $(0.1, 3, 0.5, 4)$  & $(0.1, 3, 8)$ & $(0.00001, 3, 9)$ & $(0.00001, 0.0001, 3, 9)$ \\
phyllodes\_tumour\_vs\_lobular\_carcinoma & $(1000, 143, 2)$ & $(100000, 103, 2)$ & $(0.0001, 103, 8, 3)$  & $(10000, 103, 5)$ & $(0.01, 23, 9)$ & $(10, 0.0001, 43, 1)$ \\
phyllodes\_tumour\_vs\_mucinous\_carcinoma & $(0.0001, 183, 8)$ & $(0.0001, 183, 8)$ & $(100, 163, 0.125, 5)$  & $(1000, 103, 8)$ & $(0.1, 123, 1)$ & $(0.001, 10000, 43, 8)$ \\
phyllodes\_tumour\_vs\_papillary\_carcinoma & $(1000, 123, 2)$ & $(1, 23, 9)$ & $(0.001, 103, 0.0625, 2)$  & $(10000, 103, 7)$ & $(10, 123, 9)$ & $(0.001, 100, 23, 9)$ \\
tubular\_adenoma\_vs\_ductal\_carcinoma & $(0.0001, 183, 3)$ & $(0.00001, 123, 1)$ & $(0.00001, 3, 2, 2)$  & $(0.0001, 203, 2)$ & $(0.001, 143, 2)$ & $(0.001, 0.0001, 103, 9)$ \\
tubular\_adenoma\_vs\_lobular\_carcinoma & $(0.01, 23, 1)$ & $(10, 23, 1)$ & $(0.01, 23, 8, 3)$  & $(1000, 23, 9)$ & $(1000, 43, 2)$ & $(10000, 100, 63, 9)$ \\
tubular\_adenoma\_vs\_mucinous\_carcinoma & $(0.00001, 103, 2)$ & $(0.00001, 103, 2)$ & $(0.1, 63, 16, 7)$   & $(10000, 203, 3)$ & $(0.0001, 143, 2)$ & $(100000, 0.01, 83, 1)$ \\
tubular\_adenoma\_vs\_papillary\_carcinoma & $(0.00001, 63, 2)$ & $(0.00001, 3, 1)$ & $(0.01, 143, 64, 2)$  & $(0.01, 103, 4)$ & $(10000, 163, 5)$ & $(0.001, 100000, 23, 1)$ \\ \hline
\multicolumn{7}{l}{$^{\bigstar}$ denotes the proposed models.}\\
\end{tabular}}
\end{table*}
\end{landscape}

\end{document}